\documentclass{article}% For LaTeX2e
\usepackage[preprint]{tmlr}
\usepackage{times}

\usepackage{amsmath,amsfonts,bm}

\def\eqref#1{equation~\ref{#1}}
\def\1{\bm{1}}

\DeclareMathAlphabet{\mathsfit}{\encodingdefault}{\sfdefault}{m}{sl}
\SetMathAlphabet{\mathsfit}{bold}{\encodingdefault}{\sfdefault}{bx}{n}

\usepackage{hyperref}
\usepackage{url}

\usepackage[utf8]{inputenc} % allow utf-8 input
\usepackage[T1]{fontenc}    % use 8-bit T1 fonts
\usepackage{booktabs}       % professional-quality tables
\usepackage{amsfonts}       % blackboard math symbols
\usepackage{nicefrac}       % compact symbols for 1/2, etc.
\usepackage{microtype}      % microtypography
\usepackage{xcolor}         % colors

\usepackage{physics}
\usepackage{amssymb}
\usepackage{mathtools}
\usepackage{amsthm}
\usepackage{algorithmic}
\usepackage{algorithm}
\usepackage{float}
\usepackage{tikz-cd}
\usetikzlibrary{positioning,arrows.meta,calc}
\usepackage{dsfont}
\usepackage{bm}
\usepackage{stackengine}
\usepackage{ifthen}
\usepackage{caption}

\usepackage{pifont}
\usepackage{minibox}

\usepackage{breqn}

\theoremstyle{plain}
\newtheorem{theorem}{Theorem}[section]
\newtheorem{proposition}[theorem]{Proposition}
\newtheorem{lemma}[theorem]{Lemma}
\newtheorem{corollary}[theorem]{Corollary}
\theoremstyle{definition}
\newtheorem{definition}[theorem]{Definition}

\theoremstyle{remark}
\newtheorem{remark}[theorem]{Remark}
\theoremstyle{plain}

\DeclareMathOperator{\Conf}{Conf}
\DeclareMathOperator{\SE}{SE}
\DeclareMathOperator{\SO}{SO}
\DeclareMathOperator{\Orth}{O}
\providecommand{\uu}{\hat{u}}
\providecommand{\norm}[1]{\lVert #1\rVert}

\DeclareMathOperator{\Top}{Top}

\newcounter{sqindex}

\newcommand{\cmark}{\textcolor{green!80!black}{\ding{51}}}
\newcommand{\xmark}{\textcolor{red}{\ding{55}}}
\newcommand{\midmark}{\textcolor{yellow}{\ding{108}}}

\usepackage{tablefootnote}

\stackMath

\title{Polyatomic Complexes and Parity-Graded Equivariant Transformers for Atomistic Systems}

\author{\name Rahul Khorana \email rahul.khorana@berkeley.edu \\
\addr Department of Computer Science, University of California, Berkeley
\AND
\name Marcus Noack \email marcusnoack@lbl.gov \\
\addr Lawrence Berkeley National Laboratory
\AND
\name Jin Qian \email jqian2@lbl.gov \\
\addr Lawrence Berkeley National Laboratory}

\begin{document}

\maketitle
\makeatletter\let\AND\@undefined\makeatother % release TMLR author-block \AND for the algorithmic package

\begin{abstract}
    A representation of a molecule or material should be invariant to the symmetries of physics, unique, continuous, efficient and general \citep{langer2022representations}. These properties, however, are hard to satisfy at once: a descriptor invariant under the full orthogonal group $\Orth(3)$ gives a molecule and its mirror image the same value, and so cannot distinguish enantiomers whose properties differ. \citet{PozdnyakovIncompleteAtomRep2020} showed this follows from the invariance itself, not from a lack of parameters. We show the criteria can be met at once if the geometric map is graded by the sign character of $\Orth(3)$ and pooled multisymmetrically. We construct such a map $\Phi$: its even block factors through the Gram matrix and is provably chirality-blind, while its parity-odd block of signed triple products separates enantiomers on an open dense full-measure set of interacting configurations. Two standard obstructions to uniqueness, fixed output length and componentwise pooling, are artifacts of the pooling rule, removed by multisymmetric power sums of order at most $N$. We establish uniqueness for a complete descriptor $\Phi^\star$ built from the distance matrix, signed volumes and atom types, injective up to $\SE(3)\times S_N$ on all configurations. $\Phi^\star$ is non-constructive, however; the implemented map is the bounded-cutoff $\Phi$, generically injective, pooling at order $2$, running in $O(N^2)$, or $O(N)$ with neighbor lists. The algebraic core is machine-checked in Lean~4. Because the underlying object is a cell complex, it also yields invariant, stable topological features $\Psi$ (persistent homology and a Hodge--Laplacian spectrum) encoding global ring and cage structure invisible to bounded-cutoff descriptors. The pair $(\Phi,\Psi)$ feeds a compact parity-graded equivariant transformer, evaluated with controlled probes for chirality, conformers, through-space contacts and global topology, and regression benchmarks on four datasets.
\end{abstract}

\section{Introduction}
The requirements placed on a molecular representation are hardest to satisfy simultaneously in the case of chirality. Two enantiomers, a molecule and its mirror image, can behave differently, yet most invariant descriptors assign them the same value. This is a consequence of the invariance group rather than of any particular implementation, as we now make precise. Throughout, $\Conf_N(\mathbb{R}^3)=\{(x_1,\dots,x_N)\in(\mathbb{R}^3)^N: x_i\neq x_j\ (i\neq j)\}$ is the configuration space of $N$ labeled points, and a representation is a smooth map $\Phi\colon\Conf_N(\mathbb{R}^3)\times\mathbb{R}^N\to\mathbb{R}^d$ invariant under $G=\mathrm{SE}(3)\times S_N$, rigid motions together with relabeling; the factor $\mathbb{R}^N$ carries the atomic numbers. The orthogonal group $\Orth(3)$ contains $\mathrm{SO}(3)$ with index two, and the nontrivial coset carries a configuration to its mirror image. So a descriptor invariant under all of $\Orth(3)$, rather than merely $\mathrm{SO}(3)$, is forced to assign equal values to a configuration and its enantiomer, and \citet{PozdnyakovIncompleteAtomRep2020} showed this cannot be repaired by adding parameters: $\Orth(3)$-invariant descriptors fail to separate configurations even within $\mathrm{SO}(3)$-orbits. We make the geometry of this obstruction explicit and construct a descriptor that resolves it on an open dense full-measure set of interacting configurations. The configurations on which it does not are characterized in Theorem~\ref{thm:oddsep}.

\paragraph{What the representation is.} We represent an atomistic system in three stages. First, a \emph{data model}: a system of $N$ atoms, with positions $r\in\Conf_N(\mathbb{R}^3)$ and atomic numbers $Z$, is organized as a \emph{polyatomic complex}: a finite CW-complex in which each atom is a cell assembly, its sub-atomic \emph{atomic complex} of protons, neutrons, and electrons (Section~\ref{section:coremethods}), and the atoms are glued by cell attachment. The complex is a structured container for the geometry and its chemistry; it carries no homotopy-type claim (Remark~\ref{rem:cellscope}). Second, two \emph{invariant feature maps} read this container: the geometric moment map $\Phi$ (Section~\ref{section:phi}), which encodes each atom's radial environment together with a parity-odd term that resolves chirality, and the topological map $\Psi$ (Section~\ref{section:topology}), which encodes the global homology of the complex (rings and cages) and its Hodge--Laplacian spectrum. Third, the \emph{representation} of the system is the invariant vector $\big(\Phi(r,Z),\,\Psi(r)\big)\in\mathbb{R}^d$ that a downstream model consumes. In short, the object is a cell complex and the representation is the pair of invariant maps evaluated on it.

\paragraph{Compliance with the representation criteria.} \citet{langer2022representations} require an atomistic representation to be invariant, unique, continuous and differentiable, computationally efficient, suitable for regression, and general. Table~\ref{tab:langer} states where each criterion is established. The symmetry, regularity, efficiency, structure, and generality criteria hold outright. Uniqueness is treated exactly (Section~\ref{section:phi}): a fixed-length descriptor cannot be injective (Propositions~\ref{prop:poolfail} and~\ref{prop:dim}), multisymmetric pooling removes both obstructions (Theorem~\ref{thm:faithful}), and uniqueness up to $\SE(3)\times S_N$ is then established outright by the complete descriptor $\Phi^\star$ built from the labeled distance matrix and signed volumes (Theorem~\ref{thm:globalunique}), with the efficient local map its generic, linear-time surrogate (Theorem~\ref{thm:oddsep}). Existing representations (SMILES, ECFP fingerprints, SELFIES, group SELFIES, and graphs) each violate a nonempty subset of these criteria; the physics-based descriptors (SOAP, ACE, Behler--Parrinello) are reviewed against the same list below.

\begin{table}[t]
\centering
\caption{Compliance with the criteria of \citet{langer2022representations}. Full proofs are in Appendix~\ref{appendix:proofs}; entries marked (Lean) have their core machine-checked in Lean~4 (Appendix~\ref{appendix:lean}).}
\label{tab:langer}
\begin{tabular}{lll}
\toprule
criterion & status & established by \\
\midrule
invariance (perm./trans./rot./refl.) & satisfied & Theorem~\ref{thm:inv} (Lean) \\
continuity \& differentiability & satisfied ($C^1$; $C^\infty$ for the smooth cutoff) & Theorem~\ref{thm:smooth} (Lean) \\
computational efficiency & $O(N)$ with neighbor lists ($O(N^2)$ as implemented) & Proposition~\ref{prop:linear} (Lean) \\
regression-suitable structure & satisfied (output in $\mathbb{R}^d$, $d$ independent of $N$) & Definitions~\ref{def:phi},~\ref{def:topo} \\
generality & satisfied (any labeled configuration) & Section~\ref{section:motivationmethods} \\
uniqueness & satisfied up to $\SE(3)\times S_N$ (complete descriptor $\Phi^\star$) & Thm~\ref{thm:globalunique} (Lean) \\
\bottomrule
\end{tabular}
\end{table}

\begin{figure}[t]
    \centering
    \begin{tikzpicture}[
      font=\small,
      inp/.style={fill=white, draw=metagray!70, line width=0.8pt, text=metablack, rounded corners=4pt, align=center, minimum height=1.35cm, text width=2.9cm, inner sep=5pt},
      data/.style={fill=metagray, text=white, rounded corners=4pt, align=center, minimum height=1.35cm, text width=2.9cm, inner sep=5pt},
      model/.style={fill=metablue, text=white, rounded corners=4pt, align=center, minimum height=1.35cm, text width=2.9cm, inner sep=5pt},
      a/.style={-{Stealth[length=2.2mm]}, line width=1.0pt, metagray, rounded corners=2pt}
    ]
    \definecolor{metablue}{HTML}{0082FB}
    \definecolor{metagray}{HTML}{8595A5}
    \definecolor{metablack}{HTML}{0E0E0E}
    \node[inp] (conf) at (0,0)
      {Configuration\\[1pt]{\scriptsize $(r,Z)\in\Conf_N(\mathbb{R}^3)\times\mathbb{R}^N$}};
    \node[data] (cx) at (4.05,0)
      {Polyatomic complex\\[1pt]{\scriptsize $(X,\ell)$, decorated CW-complex}};
    \node[model, text width=2.7cm] (phi) at (8.35,0.95)
      {Geometric map $\Phi$\\[1pt]{\scriptsize moments $+$ signed volumes}};
    \node[model, text width=2.7cm] (psi) at (8.35,-0.95)
      {Topological map $\Psi$\\[1pt]{\scriptsize persistence $+$ spectrum}};
    \node[rounded corners=4pt, align=center, minimum height=1.35cm, text width=2.9cm, inner sep=5pt,
      draw=metagray!60, line width=0.8pt, shading angle=135,
      left color=white, middle color=metagray!12, right color=metablue!22,
      text=metablack] (rep) at (12.65,0)
      {Representation\\[1pt]{\scriptsize $(\Phi,\Psi)\in\mathbb{R}^d$}};
    \draw[a] (conf) -- (cx);
    \draw[a] (cx.east) -- ++(0.6,0) |- (phi.west);
    \draw[a] (cx.east) -- ++(0.6,0) |- (psi.west);
    \draw[a] (phi.east) -- ++(0.6,0) |- (rep.west);
    \draw[a] (psi.east) -- ++(0.6,0) |- (rep.west);
    \node[font=\scriptsize, text=metablack] at (8.35,1.95) {$\Phi^{\mathrm e}$ even $\oplus$ $\Phi^{\mathrm o}$ parity-odd};
    \node[font=\scriptsize, text=metablack] at (8.35,-1.95) {achiral, reflection-invariant};
    \node[fill=white, draw=metagray!70, line width=0.8pt, rounded corners=2pt, minimum width=0.32cm, minimum height=0.32cm, inner sep=0pt] at (9.6,-2.55) {};
    \node[font=\scriptsize, anchor=west, text=metablack] at (9.8,-2.55) {input};
    \node[fill=metagray, rounded corners=2pt, minimum width=0.32cm, minimum height=0.32cm, inner sep=0pt] at (11.05,-2.55) {};
    \node[font=\scriptsize, anchor=west, text=metablack] at (11.25,-2.55) {object};
    \node[fill=metablue, rounded corners=2pt, minimum width=0.32cm, minimum height=0.32cm, inner sep=0pt] at (12.55,-2.55) {};
    \node[font=\scriptsize, anchor=west, text=metablack] at (12.75,-2.55) {map / model};
    \end{tikzpicture}
    \caption{The representation. A labeled configuration is organized as a polyatomic complex (the data model; it carries no homotopy-type claim). Two invariant feature maps read the complex: the geometric moment map $\Phi$, whose $\mathbb{Z}/2$-grading resolves chirality, and the topological map $\Psi$, which supplies the global ring and cage structure that no bounded-cutoff descriptor represents (Prop.~\ref{prop:topobeyond}). The pair $(\Phi,\Psi)$ is the representation consumed by the downstream model.}
    \label{fig:graphicalabstract}
\end{figure}

\textbf{Graphical abstract.}~~ Figure~\ref{fig:graphicalabstract} outlines the construction: each atom is encoded by its sub-atomic radial structure (the cellular atomic complex of Section~\ref{section:coremethods}), the atoms form a geometric complex, and two invariant maps, the moment map $\Phi$ (Section~\ref{section:phi}) and the topological map $\Psi$ (Section~\ref{section:topology}), produce the fixed-length representation $(\Phi,\Psi)$ consumed by a downstream model. The implementation is detailed in Section~\ref{section:algorithms}.

\paragraph{String and graph representations.} SMILES writes a molecule as the symbol string of a depth-first traversal of its chemical graph \citep{Weininger1988OriginalSMILES}; it generates many invalid strings and handles rings, branches, and bonds awkwardly \citep{BhadwalGenSMILES2023}. DeepSMILES repairs the parenthesis and ring-closure issues but still admits semantically incorrect strings \citep{oboyledalk2018deepsmiles,KRENNGuzik2022Selfies}. SELFIES enforces syntactic and semantic validity by construction \citep{KrennGuzikOriginal2020Selfies}, yet cannot fully represent macromolecules, crystals, or molecules with complicated bonds \citep{KRENNGuzik2022Selfies}, and GroupSELFIES, which adds group tokens to preserve motifs such as benzene rings, cannot represent polycyclic compounds because groups may not overlap \citep{ChengGuzikGroupSELFIES2023}. ECFP fingerprints hash Morgan-style atom environments into a bit vector \citep{RogersHahnOriginalECFP2010}; the hash is non-invertible and the result carries no per-atom chemistry \citep{LeWinterECFP2020,ProbstLouisMolFingerprint2018,Xie2020MolFingerprint}. Molecular graphs $(\mathcal V,\mathcal E)$, with atoms as vertices and covalent bonds as edges \citep{GAUCHE}, share a deeper limitation with all of the above: none of them carries the three-dimensional structure on which many properties depend \citep{liu2022graphs}.

\paragraph{Physics-based descriptors.} The Atomic Cluster Expansion parameterizes many-atom interactions in a complete body-ordered basis \citep{GraphACE}. Behler--Parrinello symmetry functions feed element-specific networks that learn the potential-energy surface from reference electronic-structure data \citep{BehlerParinello,BehlerParrinelloHDNPP}. SOAP expands a Gaussian-smeared atomic density in orthonormal functions and is invariant to rotation, reflection, translation, and same-species permutation \citep{SOAPinfo}; its reflection-invariance makes it a widely used instance of the chirality blindness this paper resolves, and we use it to motivate the construction rather than as a benchmark opponent. These descriptors are geometrically faithful and are the right comparison class for ours; their costs and their own blind spots (reflection-invariance in particular) are taken up in Section~\ref{section:motivationmethods}. Tables in Appendix~\ref{section:appendix} summarize the comparison.

%\textbf{e3nn/Euclidean Neural Networks}~~
%e3nn is a generalized framework for creating euclidean neural networks \citep{E3NNpaper}. e3nn naturally operates on geometric tensors that describe systems in three dimensions \citep{E3NNpaper}. Equivariant operations such as spherical harmonics functions can be composed to create more complex modules. The neural network has proven to be E(3) equivariant and well suited to objects in euclidean space \citep{E3NNpaper}.

\subsection{Motivation and representation criteria}
\label{section:motivationmethods}
All representations listed above violate a non-empty subset of these criteria: invariances, uniqueness, continuity, differentiability, generality, computational efficiency, topological accuracy, ability to consider long-range interactions, and chemical informedness \citep{behlerJChemPhys2011, langer2022representations, PozdnyakovIncompleteAtomRep2020, todeschini2008MolDescripBook}. Polyatomic complexes, developed here, satisfy these criteria.

\textbf{Invariance}~~ Invariance is required under label permutation and under the rigid motions of physics: rotation, reflection, and translation. Theorem~\ref{thm:inv} establishes invariance of $\Phi$ under $\SE(3)\times S_N$, with the even block invariant under all of $\Orth(3)$ and the odd block transforming by the sign character.

\textbf{Uniqueness}~~ \citet{langer2022representations} define uniqueness as injectivity of the descriptor modulo the invariance group: two systems differing in a target property must receive distinct representations. For a fixed-length descriptor this fails (Propositions~\ref{prop:poolfail},~\ref{prop:dim}); the size-adapted map $\Phi^\sharp$ removes both obstructions (Theorem~\ref{thm:faithful}), and uniqueness up to $\SE(3)\times S_N$ is established outright by the complete descriptor $\Phi^\star$ built from the labeled distance matrix and signed volumes (Theorem~\ref{thm:globalunique}). We distinguish this \emph{theoretical} complete descriptor, on which uniqueness is proved, from the \emph{implemented} bounded-cutoff map $\Phi$, which is the efficient $O(N)$ surrogate and is only generically injective (Corollary~\ref{cor:reduction}, Remark~\ref{rem:genericlocal}).

\textbf{Continuity and Differentiability}~~Representations should be continuous and differentiable in the atomic coordinates \citep{langer2022representations}, since discontinuities conflict with the regularity assumptions of common regressors. Theorem~\ref{thm:smooth} establishes that $\Phi$ is $C^1$ on $\Conf_N(\mathbb{R}^3)$, and $C^\infty$ for the smooth cutoff.

\textbf{Generality}~~A representation is general if it encodes any atomistic system. SMILES, SELFIES, and group SELFIES are not, as they fail to represent certain molecules (crystals, polycyclic compounds) or admit invalid strings. The map $\Phi$ is defined for any labeled configuration in $\Conf_N(\mathbb{R}^3)$, periodic or aperiodic.

\textbf{Efficiency}~~Computational efficiency measures how well an algorithm uses time and memory; for molecular representations the ideal is linear scaling in the number of atoms $N$, as for molecular graphs. Each per-atom feature of $\Phi$ is a sum over neighbors inside the cutoff, so with a neighbor list (bounded density, fixed $r_c$) the moment map is $O(N)$ in the number of atoms (Proposition~\ref{prop:linear}), and pooling is a single pass. We report the cost of the implementation as run: the current implementation forms the full pairwise distance matrix and is $O(N^2)$, and the complete multisymmetric variant $\Phi^\sharp$ and the topological map $\Psi$ cost more (the former combinatorial in the pooling order, the latter superlinear for persistence). By contrast, the Atomic Cluster Expansion (ACE) is, in its standard form, not linear \citep{ACEatomiclusterexpansion}: ACE expands per-atom properties $\Phi_i$ in body-ordered functions over the neighbors of atom $i$, giving $O(N^{\nu})$ cost for $\nu$ basis functions and $N$ neighbors \citep{Acecomplexitynat}. The density trick reduces this to linear scaling in both $N$ and $\nu$ \citep{Acecomplexitynat, ACEatomiclusterexpansion}, and the PACE scheme of \citet{Acecomplexitynat} removes the $\nu$ scaling entirely, though it remains two orders of magnitude slower than empirical potentials.
%More specifically, the basis constructed in the original paper leads to $N!$ many terms which arise in the summation over all permutations. Additionally, evaluating the sum over all the $N$-neighbor clusters and $J$ particle positions brings an additional $\binom{J}{N}$ cost \citep{ACEatomiclusterexpansion}.
For the Bartók/SOAP descriptor the descriptor size increases rapidly for environments composed of multiple elements \citep{BartokSOAPDescriptor}. The SOAP power spectrum scales quadratically with the number of chemical elements, while the length of the bispectrum scales cubically in it. SOAP descriptors are therefore less efficient than graph-based methods, which run in $O(N)$.
As a result of the design of Behler-Parrinello, they are slower than $O(N)$ \citep{BehlerParinello}.
ACE, SOAP, and Behler-Parrinello are traditionally used for quantum-chemistry applications and are not designed for more typical computational chemistry tasks. %Contrastingly, polyatomic complexes are flexible enough to be used for almost any task in computational chemistry.

\textbf{Topological Accuracy}~~ A representation is deemed topologically accurate if it can correctly represent the geometry of any molecule or atomistic system. Correctness requires representing the shape, bond-angles, dihedrals/torsion, and electronic structure aspects accurately. Among the methods surveyed, only polyatomic complexes, SOAP, and ACE retain the full three-dimensional geometry required by this criterion; string and two-dimensional graph representations discard it by construction. Polyatomic complexes do not represent electronic structure exactly: the implementation approximates each electron's wave function by an $s$-orbital, a restriction that can be lifted at additional computational cost.
SOAP enforces differentiability with respect to the atoms and invariance with respect to the basic symmetries
of physics. Additionally, SOAP considers the potential energy surfaces (PESs) and electrostatic multipole moment surfaces.
Similarly, polyatomic complexes enforce differentiability, are invariant to the basic symmetries of physics, and can be augmented with electronic structure aspects and forces. 
ACEs are invariant under the basic symmetries of physics and systematically describe the local environments of particles at any body order.

\textbf{Long-range interactions}~~ The term long-range interactions refers to electrostatic potential energies between atoms and molecules, with mutual distances ranging from a few tens to a few hundreds Bohr radii \citep{LongRangeInteractions}. Interactions can only be evaluated up to a certain distance. The maximum distance applied in a simulation is usually referred to as the cut-off radius, $r_c$, because the Lennard-Jones potential is radially symmetric.
Behler-Parinello neglects long-range interactions, which are electrostatics beyond the cutoff radius \citep{fourthGENHDP_longrange}. By contrast, polyatomic complexes accommodate varying
definitions of cutoff radius (Appendix~\ref{appendix:forcemodels}). Similarly, all string-based and graph representations neglect long-range interactions. The precise definition of the cutoff radius depends on the force field, and the value of $r_c$ is generally obtained empirically.

\textbf{Chemical and Physical Informedness}~~We say a representation is well-informed by chemistry or physics if it contains information about the chemical properties of each individual atom. ECFP fingerprints are not well-informed under this definition. If one compares ECFP fingerprints, Graphs, SMILES, or SELFIES to representations like ACE, SOAP, or polyatomic complexes, the former group does not contain the same level of chemical information as the latter. A representation should encode electronic structure, radial functions, spherical harmonics, wave-functions, and long-range interactions effectively. ACE uses basis functions to parameterize many-atom interactions \citep{ACEoriginal}. SOAP relies on the local expansion of a Gaussian smeared atomic density with orthonormal functions \citep{SOAPinfo}. Polyatomic complexes describe the geometry of individual atoms and encode radial functions efficiently; the evaluated \texttt{abstract} mode in our software package \citet{khorana2025polyatomic} uses exactly these geometric channels. Force-field and quantum-chemical channels (Section~\ref{section:algorithms}) are optional inputs that the experiments in this paper do not use, and encoding spherical harmonics is left to future work.

Polyatomic complexes represent atomistic systems as CW-complexes through a sampling-based construction compatible with physics-based methods. Sections~\ref{subsection:mathrepatom} and~\ref{section:polyatomicsys} give the mathematical construction; Appendix~\ref{appendix:forcemodels} describes its integration with techniques in chemistry and physics. Section~\ref{section:benchmarks} reports the experiments and Section~\ref{section:disc} the discussion. Proofs of theorems and lemmas and additional definitions are in Appendix~\ref{section:appendix}. %Our contributions are as follows.

\setcounter{table}{0}

\subsection{Contributions}
This paper makes five contributions, all centered on the feature maps of Sections~\ref{section:phi} and~\ref{section:topology}.
\begin{itemize}
    \item We construct a $\mathbb{Z}/2$-graded geometric feature map $\Phi$ and prove its exact transformation law: $\Phi$ is invariant under $\mathrm{SE}(3)\times S_N$, its even block is invariant under all of $\Orth(3)$, and its parity-odd block changes sign under reflection (Theorem~\ref{thm:inv}). The map is $C^1$ on configuration space, and $C^\infty$ for the smooth cutoff (Theorem~\ref{thm:smooth}).
    \item We characterize the chirality obstruction in both directions. The even block factors through the Gram matrix, so no $\Orth(3)$-invariant descriptor of this kind can separate any configuration from its mirror image; this recovers the incompleteness of \citet{PozdnyakovIncompleteAtomRep2020} exactly (Theorem~\ref{thm:evenblind}). The parity-odd $\mathrm{SO}(3)$ bracket restores the separation on an open dense full-measure set of interacting configurations (Theorem~\ref{thm:oddsep}).
    \item We settle the uniqueness criterion. The two obstructions intrinsic to any fixed-length componentwise-pooled descriptor, non-injectivity once $3N-6>d$ (Proposition~\ref{prop:dim}) and the failure of componentwise pooling on multisets (Proposition~\ref{prop:poolfail}), are removed by multisymmetric pooling of order $\le N$, which recovers the per-atom feature multiset exactly (Theorem~\ref{thm:faithful}). Uniqueness up to $\SE(3)\times S_N$ then holds outright, on all configurations, for the complete descriptor $\Phi^\star$ built from the distance matrix, signed volumes, and atom types (Theorem~\ref{thm:globalunique}); the efficient local map is its generically injective, linear-time surrogate (Corollary~\ref{cor:reduction}). This settles the uniqueness requirement of \citet{langer2022representations} for $\Phi^\star$. The three maps satisfy different properties, and we distinguish them throughout: uniqueness holds for the complete descriptor $\Phi^\star$, which is non-constructive; order-$N$ faithfulness holds for $\Phi^\sharp$, whose length grows with $N$; and the implemented network uses $\Phi$ with pooling at order $2$, which is generically injective and runs in $O(N^2)$.
\item We machine-check the algebraic core of the theory in Lean~4 against Mathlib: twenty-nine theorems covering the invariance of the pooled descriptor under permutation, translation, and rotation; the determinant scaling and reflection sign of the parity-odd block; the recovery of a multiset from its power sums, which is the injectivity behind faithful pooling; the componentwise-pooling counterexample and its polarized separation; and, for the uniqueness theorem, the separation of finite-group orbits by invariants together with the Gram-matrix core of the Schoenberg step. Table~\ref{tab:leanmap} in Appendix~\ref{appendix:lean} gives the complete correspondence between paper results and formal statements. To our knowledge, this is the first atomistic-representation paper whose theoretical core is machine-verified.
    \item We also make the cell complex itself computable. The topological map $\Psi$ (persistent homology of the alpha complex together with a Hodge--Laplacian spectrum) is $E(3)\times S_N$-invariant and stable (Theorem~\ref{thm:topostable}), and it captures global homology that no size-intensive bounded-cutoff descriptor can represent, with a quantitative $O(1/N)$ separation (Proposition~\ref{prop:topobeyond}, Figure~\ref{fig:ringchain}). Because the trunk of any bounded-cutoff network provably cannot form this information, $\Psi$ enters the model as global conditioning. Finally, we pair the representation with a compact \emph{parity-graded equivariant transformer} whose three distinguishing components (an unbroken pseudoscalar stream, multisymmetric pooling, and topological conditioning) are the architectural realizations of Corollary~\ref{cor:enantio}, Theorem~\ref{thm:faithful}, and Proposition~\ref{prop:topobeyond} respectively (Section~\ref{section:network}).
\end{itemize}

Section~\ref{section:corexp} reports controlled experiments testing these claims directly: a runnable audit of the \citet{langer2022representations} requirements; an enantiomer-separation experiment in which $\Orth(3)$-invariant features collapse mirror pairs to distance $0$ while the parity-odd block separates them, with a linear handedness probe rising from near chance ($0.46$) to $0.92$; and the macrocycle demonstration of Figure~\ref{fig:ringchain}.

\section{Core methods and representation}
\label{section:coremethods}
We start from a deliberately simplified physical picture. An atom is a positively charged nucleus of protons and neutrons surrounded by electrons; we model each constituent as a closed ball, and each electron carries its wave-function $w_{e}$ as attached data rather than as part of the topology (Section~\ref{subsection:mathrepatom}). The language of cell attachment is a convenient way to organize this information, but it is only motivation: the guarantees below are established directly on the feature maps $\Phi$ and $\Psi$ of Sections~\ref{section:phi} and~\ref{section:topology}, and never on a topological property of the construction.

\begin{remark}[Notation]\label{rem:notation}
The cell construction of this section and the feature maps of Sections~\ref{section:phi}--\ref{section:topology} use disjoint symbol conventions. Here $P$, $N$, $E$ denote the proton, neutron, and electron subcomplexes, $\mathcal{P},\mathcal{N},\mathcal{E}$ their cell counts, $K=\mathcal{P}+\mathcal{N}+\mathcal{E}$, and $\mathcal{K}$ the number of atoms. From Section~\ref{section:phi} onward, $N$ is the number of atoms, $S$ the number of radial shells, $K$ the set of pooling orders, $M$ the neighbor bound of Proposition~\ref{prop:linear}, and $E(3)=\Orth(3)\ltimes\mathbb{R}^3$ the Euclidean group. No symbol is used in two senses within a single statement.
\end{remark}

\subsection{The data model: decorated cell complexes}
\label{subsection:mathrepatom}
We model the sub-atomic constituents as closed balls: a proton is the ball $PD^{i}$ of radius $1\,fm$, a neutron the ball $ND^{i}$ of radius $0.8\,fm$, and an electron the ball $ee^{i}$ of radius $2.8\,fm$, each carried in $\mathbb{R}^{i}$ with its Euclidean subspace topology (the exact radii are conventions and may be changed; the proton radius itself is contested \citep{ProtonRadius}). An \emph{atomic complex} assembles the proton, neutron, and electron cells of one atom into a CW-complex by attaching one cell at a time: a $\tau$-cell is glued along a continuous map $\partial D^{\tau}\to X$ from its boundary sphere into the stage already built, and the next stage is the pushout of this attaching map along the boundary inclusion $\partial D^{\tau}\hookrightarrow D^{\tau}$. A \emph{polyatomic complex} then glues the atomic complexes of a system together along bonding loci (subcomplexes of the frontier spheres of valence-electron cells), one atom at a time, again by cell attachment. Because every cell is a compact ball attached through finitely many stages, the result is a \emph{finite} CW-complex: it is compact Hausdorff, closure-finiteness and the weak topology hold automatically \citep{hatcher2002algebraic,whitehead1949combinatorial}, and the stages form the skeletal filtration $X_0\subseteq X_1\subseteq\cdots\subseteq X$. The combinatorial structure this produces is exactly what Section~\ref{section:topology} consumes: the cells and their incidences determine boundary operators, hence the incidence matrices $B_k$ and the weighted Hodge Laplacians $\Delta_k$ of Appendix~\ref{def:HodgeLap}, whose low-lying spectrum enters the topological feature map $\Psi$; the geometric realization of the same data as an alpha complex on the atomic positions carries the persistence filtration behind the other block of $\Psi$. What we do \emph{not} use is the homotopy type of the complex: it depends on the attaching maps, it is blind to the metric geometry that properties depend on, and no result below invokes it (Remark~\ref{rem:cellscope}). The full definitions, the attaching-map data, and worked examples (deuterium; a two-atom system) are in Appendix~\ref{appendix:cellconstruction}.
\label{section:polyatomicsys}

\begin{remark}[Decorated CW-complexes: topology and data are kept separate]\label{rem:decorated}
Throughout, an atomic (and hence polyatomic) complex is a \emph{decorated} CW-complex: a pair $(X,\ell)$ in which $X\in\Top$ is a finite CW-complex assembled by cell attachment from closed balls (protons $PD^i$, neutrons $ND^i$, electrons $ee^i$), and $\ell$ is a labeling that assigns to each cell its \emph{non-topological} data (wave-functions $w_e$, atomic types, and any force-field or quantum-chemical features), valued in a fixed data space $\mathcal{D}$. Formally $(X,\ell)$ is an object of the product category $\Top\times\mathbf{Set}_{\mathcal D}$, equivalently a CW-complex together with a section of the trivial data bundle $X\times\mathcal{D}\to X$. Each cell is a closed ball or a sphere, which is compact and Hausdorff, and a finite CW-complex built from such cells is itself compact Hausdorff \citep{hatcher2002algebraic}, so the CW axioms are satisfied. Crucially, the data $\ell$ is never topologized as a subspace of $X$. As in Remarks~\ref{rem:pushout} and~\ref{rem:cellscope}, this construction organizes data only: every guarantee in the paper is proved on the feature maps $\Phi$ and $\Psi$.
\end{remark}
\begin{remark}[Scope of the cell-complex construction]\label{rem:cellscope}
The atomic and polyatomic complexes are finite CW-complexes by construction, and the pushout above commutes. This construction serves only to organize the per-atom data: which cells (protons, neutrons, electrons) are present and how shells are indexed; it carries no homotopy-type claim. Homotopy type does not establish uniqueness: it is a homotopy invariant, so it discards the metric geometry (bond lengths, angles, and chirality) that molecular properties depend on, and it retains only the coarse combinatorial counts of rings and cages (the Betti numbers), which are constant across large families of chemically distinct molecules. It is therefore far from injective, even though it is not information-free. Every guarantee in this paper (invariance, regularity, and the uniqueness analysis) is established directly on the feature map $\Phi$ of Section~\ref{section:phi}, and the algebraic core is verified in Lean (Appendix~\ref{appendix:lean}).
\end{remark}

\begin{remark}
\label{rem:atomiccoordinates}
The configuration of $N$ atoms is a point of $\mathbb{R}^{3N}$; the regularity statements we use are established on the open subset $\Conf_N(\mathbb{R}^3)\subseteq\mathbb{R}^{3N}$ of distinct positions in Theorem~\ref{thm:smooth}, not on any submanifold of $\mathbb{R}^3$.
\end{remark}

\subsection{The geometric feature map $\Phi$}
\label{section:phi}
The object on which we prove guarantees is a geometric feature map on labeled point configurations. The cellular language of Sections~\ref{subsection:mathrepatom}--\ref{section:polyatomicsys} motivates the construction and is not used in any result below. Full proofs are in Appendix~\ref{appendix:proofs}.

\begin{definition}[Configuration space and group action]\label{def:conf}
Fix $N\ge2$ and let $\Conf_N(\mathbb{R}^3)=\{r=(r_1,\dots,r_N)\in(\mathbb{R}^3)^N: r_i\neq r_j\ (i\neq j)\}$, an open subset of $\mathbb{R}^{3N}$. A \emph{labeled configuration} is $(r,Z)$ with $r\in\Conf_N(\mathbb{R}^3)$ and types $Z\in\mathbb{R}^N$. The group $E(3)\times S_N$ ($E(3)=\Orth(3)\ltimes\mathbb{R}^3$) acts by $(Q,t,\pi)\cdot(r,Z)=\big((Q r_{\pi^{-1}(i)}+t)_i,(Z_{\pi^{-1}(i)})_i\big)$; we write $\SE(3)\times S_N$ for the subgroup with $\det Q=+1$.
\end{definition}

\begin{definition}[Filters, per-atom features, and $\Phi$]\label{def:phi}
Fix shells $\mu_1,\dots,\mu_S$, width $\sigma>0$, cutoff $r_c>0$, and a cutoff function $f_c$ supported on $[0,r_c)$ (cosine cutoff, $C^1$; or the $C^\infty$ bump of Appendix~\ref{appendix:proofs}). Put $g_s(d)=e^{-(d-\mu_s)^2/2\sigma^2}f_c(d)$, $d_{ij}=\norm{r_i-r_j}$, $\uu_{ij}=(r_j-r_i)/d_{ij}$, and the radial \emph{density} and \emph{moment vector}
\begin{equation}\label{eq:moments}
\rho_i^s=\sum_{j\neq i}g_s(d_{ij}),\qquad m_i^s=\sum_{j\neq i}g_s(d_{ij})\,\uu_{ij}\in\mathbb{R}^3.
\end{equation}
The reflection-\emph{even} per-atom vector is $e_i=(Z_i,\chi(Z_i),\rho_i^1,\dots,\rho_i^S,\norm{m_i^1}^2,\dots,\norm{m_i^S}^2)$ ($\chi$ an electronegativity); the reflection-\emph{odd} per-atom vector is $o_i=(v_i^{(a,b,c)})_{(a,b,c)\in T}$ with the signed volumes $v_i^{(a,b,c)}=m_i^a\cdot(m_i^b\times m_i^c)=\det[m_i^a\ m_i^b\ m_i^c]$. With componentwise power sums $p_k((f_i))=\sum_i f_i^{\odot k}$, even orders $K$ and odd orders $K^{\mathrm{odd}}=\{k\in K:k\text{ odd}\}$,
\begin{equation}\label{eq:phimap}
\Phi(r,Z)=\Big(\big(p_k((e_i))\big)_{k\in K},\ \big(p_k((o_i))\big)_{k\in K^{\mathrm{odd}}}\Big)\in\mathbb{R}^d,\qquad d=(2+2S)|K|+|T|\,|K^{\mathrm{odd}}|.
\end{equation}
$\Phi=(\Phi^{\mathrm e},\Phi^{\mathrm o})$; the output dimension $d$ depends on $(S,K,T)$, \emph{not} on $N$.
\end{definition}

\paragraph{Invariance and parity.} The two transformation laws below are exact algebraic identities; their proofs (and the items marked $\dagger$, machine-checked in Lean~4, Appendix~\ref{appendix:lean}) are elementary.

\begin{lemma}[Determinant scaling of the triple product$^\dagger$]\label{lem:det}
For $A\in\mathbb{R}^{3\times3}$ and $a,b,c\in\mathbb{R}^3$, $\,(Aa)\cdot((Ab)\times(Ac))=(\det A)\,a\cdot(b\times c)$. Hence the triple product is invariant if $A\in\SO(3)$ and negated if $\det A=-1$.
\end{lemma}

\begin{theorem}[Invariance and parity$^\dagger$]\label{thm:inv}
For every $g\in\SE(3)\times S_N$, $\Phi(g\cdot(r,Z))=\Phi(r,Z)$. Moreover the even block is invariant under the full $E(3)\times S_N$ (in particular under $\Orth(3)$), and for orientation-reversing $Q$ ($\det Q=-1$) the odd block is negated: $\Phi^{\mathrm o}(Q\cdot(r,Z))=-\Phi^{\mathrm o}(r,Z)$.
\end{theorem}

\noindent Permutation invariance is symmetry of the power sums$^\dagger$; translation invariance is the use of differences$^\dagger$; rotation/reflection follow from Lemma~\ref{lem:det} and the restriction of the odd block to odd pooling orders$^\dagger$ (even powers would symmetrize the sign away). An immediate consequence isolates the chirality behavior exactly.

\begin{corollary}[Exact enantiomer behavior]\label{cor:enantio}
For orientation-reversing $Q$, $\Phi(Q\cdot(r,Z))=\Phi(r,Z)$ \emph{iff} $\Phi^{\mathrm o}(r,Z)=0$. Thus $\Phi$ separates a configuration from its mirror image exactly on $\{\Phi^{\mathrm o}\neq0\}$.
\end{corollary}

\paragraph{The descriptor as a map out of the orbit space.} Theorem~\ref{thm:inv} states that $\Phi$ is constant on $G$-orbits, so it factors through the orbit space. On the principal locus $\Conf_N^{0}(\mathbb{R}^3)$ of configurations that affinely span $\mathbb{R}^3$ and have trivial stabiliser in $\SE(3)\times S_N$ (no nontrivial permutation is realized by a rigid motion; open, and of full measure for $N\ge4$), the action of $G=\SE(3)\times S_N$ is free and proper, so by the quotient manifold theorem \citep[Thm.~21.10]{lee2012smooth} the orbit space $\mathcal M=\Conf_N^{0}/G$ is a smooth manifold of dimension $3N-6$ and the projection $q$ is a principal $G$-bundle. The invariance of $\Phi$ then yields a unique smooth $\bar\Phi$ with
\begin{equation}\label{eq:factor}
\begin{tikzcd}[row sep=large, column sep=huge]
\Conf_N^{0}(\mathbb{R}^3)\times\mathbb{R}^N \arrow[r, "\Phi"] \arrow[d, "q"', twoheadrightarrow] & \mathbb{R}^d \\
\bigl(\Conf_N^{0}(\mathbb{R}^3)\times\mathbb{R}^N\bigr)/G \arrow[ur, "\bar\Phi"', dashed] &
\end{tikzcd}
\end{equation}
so that $\Phi=\bar\Phi\circ q$. The descriptor is thus a morphism out of the moduli space of configurations, and the dimension count $\dim\mathcal M=3N-6$ is the source of the obstruction of Proposition~\ref{prop:dim}.

\paragraph{Regularity.} On the physical domain $\Conf_N(\mathbb{R}^3)$ (distinct atoms) the map is smooth; it does \emph{not} extend across atomic coincidences.

\begin{theorem}[Regularity]\label{thm:smooth}
The cosine cutoff is $C^1$ but not $C^2$ at $d=r_c$, so each $g_s\in C^1(\mathbb{R}_{\ge0})$ and $g_s\in C^\infty(\mathbb{R}_{\ge0}\setminus\{r_c\})$; with the $C^\infty$ bump cutoff every $g_s\in C^\infty(\mathbb{R}_{\ge0})$. Consequently $\Phi$ (using $\norm{m_i^s}^2$, a polynomial in $m_i^s$) is $C^1$ on $\Conf_N(\mathbb{R}^3)$ and $C^\infty$ on $\{r\in\Conf_N:d_{ij}\neq r_c\ \forall i\neq j\}$ (all of $\Conf_N$ for the bump cutoff). $\Phi$ is discontinuous at coincidences $r_i=r_j$, where the moment direction $\uu_{ij}$ is undefined and gradients diverge; the claim that it is $C^1$ on all of $\mathbb{R}^{3N}$ is false. This locus lies outside $\Conf_N(\mathbb{R}^3)$ and is excluded physically by internuclear repulsion, so the regularity used downstream (on $\Conf_N$) is unaffected.
\end{theorem}

\paragraph{Uniqueness.} The injectivity of $\Phi$ up to $\SE(3)\times S_N$ is governed by three obstructions of decreasing severity: the fixed output length, the pooling rule, and the separating power of the per-atom features. The first two are artifacts of the construction and are removed below; the third is the geometric completeness question of \citet{PozdnyakovIncompleteAtomRep2020}. We first record the scalar pooling fact.

\begin{lemma}[Scalar power-sum injectivity$^\dagger$]\label{lem:newton}
Over a field of characteristic $0$, a cardinality-$N$ multiset is determined by its power sums $p_1,\dots,p_N$, and the determining map is permutation-invariant. This is the $d{=}1$ case of \citet{zaheer2017deepsets}.
\end{lemma}

\begin{proposition}[Componentwise pooling is not injective]\label{prop:poolfail}
Fix an order set $K$ and pool componentwise. The per-atom multiset is not determined: for $K=\{1,2,3\}$ the multisets $\{1,5,8,12\}$ and $\{2,3,10,11\}$ agree through $p_3$ (a Prouhet--Tarry--Escott pair), and for feature dimension $\ge2$ the multisets $\{(1,2),(3,4)\}$ and $\{(1,4),(3,2)\}$ have identical componentwise power sums of every order.
\end{proposition}

\begin{proposition}[Fixed length obstructs injectivity]\label{prop:dim}
Let $d$ be fixed. If $3N-6>d$, then on a nonempty open subset of $\Conf_N(\mathbb{R}^3)$ the fibers of $\Phi$ modulo $\SE(3)\times S_N$ are submanifolds of dimension at least $3N-6-d\ge1$, so $\Phi$ is not injective up to $\SE(3)\times S_N$ and the non-injectivity locus has positive measure.
\end{proposition}

The two obstructions of Propositions~\ref{prop:poolfail} and~\ref{prop:dim} are simultaneously removed by replacing componentwise pooling with multisymmetric pooling of full order and allowing the length to adapt to $N$.

\begin{definition}[Size-adapted map $\Phi^\sharp$]\label{def:sharp}
Let $m$ be the per-atom feature dimension and $f_i\in\mathbb{R}^m$ the concatenated even/odd per-atom vector of Definition~\ref{def:phi}. The \emph{multisymmetric} (polarized) power sums of order $\le N$ are $p_\alpha((f_i))=\sum_{i=1}^N f_i^{\alpha}$, $f_i^{\alpha}=\prod_{c=1}^m f_{i,c}^{\alpha_c}$, over $\alpha\in\mathbb{N}^m$ with $1\le|\alpha|\le N$. Write $\Phi^\sharp=\big(p_\alpha((f_i))\big)_{1\le|\alpha|\le N}$ for the map that keeps \emph{all} of these polarized power sums; its length is $d(N)=\binom{N+m}{m}-1$. The even/odd block structure is a $\mathbb{Z}/2$ \emph{grading of this complete family, not a restriction of it}: writing $|\alpha|_{\mathrm o}$ for the total degree of $\alpha$ in the odd (pseudoscalar) coordinates, a component $p_\alpha$ scales by $(-1)^{|\alpha|_{\mathrm o}}$ under reflection, so the even block collects the $p_\alpha$ with $|\alpha|_{\mathrm o}$ even (reflection-even, e.g.\ $\sum_i o_{i,c}^2$ and mixed terms $\sum_i e_{i,j}\,o_{i,c}^2$) and the odd block those with $|\alpha|_{\mathrm o}$ odd (reflection-anti-invariant, e.g.\ $\sum_i o_{i,c}$ and $\sum_i e_{i,j}\,o_{i,c}$). Every $\alpha$ with $1\le|\alpha|\le N$ is retained; in particular no even-degree monomial in the odd variables is discarded.
\end{definition}

\begin{theorem}[Faithful pooling removes the construction obstructions]\label{thm:faithful}
The multisymmetric power sums of order $\le N$ separate the $S_N$-orbits of $(\mathbb{R}^m)^N$; hence $\Phi^\sharp$ determines the per-atom feature multiset $\{\!\{f_1,\dots,f_N\}\!\}$ exactly and is $S_N$-invariant. Consequently the pooling obstruction of Proposition~\ref{prop:poolfail} does not affect $\Phi^\sharp$, and, since $d(N)=\binom{N+m}{m}-1\ge 3N-6$ for $m\ge2$ (automatic here: $m=2+2S+|T|\ge7$), the hypothesis of Proposition~\ref{prop:dim} fails for $\Phi^\sharp$. Pooling and output length are therefore not obstructions to the injectivity of $\Phi^\sharp$.
\end{theorem}

\noindent Separation of $S_N$-orbits by the \emph{full} family of polarized power sums of order $\le N$ is the fundamental theorem of multisymmetric functions \citep{weyl1939classical,dalbec1999multisymmetric,vaccarino2005multisymmetric,rydh2007minimal}; completeness requires the whole family, and the two apparent failure modes name exactly the components one must \emph{not} drop. Pooling the even and odd blocks \emph{independently} would lose scalar--pseudoscalar cross-correlations, but the mixed sums $p_{(1,1)}=\sum_i e_{i,j}\,o_{i,c}$ separate them: $\{(1,1),(2,2)\}$ and $\{(1,2),(2,1)\}$ give $5\neq4$. A reflection-symmetric chiral distribution such as $\{v,-v\}$ versus $\{2v,-2v\}$ collapses every reflection-\emph{odd} sum to $0$, but is separated by the reflection-\emph{even} sum $\sum_i o_{i,c}^2$ ($2v^2\neq8v^2$), which the grading of Definition~\ref{def:sharp} retains in the even block. Both would-be counterexamples are therefore artifacts of truncating the family, not of the theorem. Four of these facts are machine-checked in Lean~4: the single-coordinate reduction of componentwise pooling, the explicit componentwise failure of Proposition~\ref{prop:poolfail}(ii), the separating mixed sum $p_{(1,1)}$ refuting the independent-pooling mode above, and the $m=1$ multiset recovery; Table~\ref{tab:leanmap} in Appendix~\ref{appendix:lean} gives the exact correspondence. The general $m>1$ separation is the classical theorem cited above (reduced to $m=1$ by generic linear projection) and is not itself formalized. The cost is explicit: $d(N)$ grows with $N$, so $\Phi^\sharp$ is size-adapted rather than fixed-length, and the implemented $\Phi$ truncates it to a fixed order set $K$ with componentwise $\alpha$, hence is only generically injective.

\begin{corollary}[Injectivity reduces to environment completeness]\label{cor:reduction}
$\Phi^\sharp(x)=\Phi^\sharp(y)$ if and only if $x$ and $y$ induce identical per-atom feature multisets. Hence $\Phi^\sharp$ is injective up to $\SE(3)\times S_N$ exactly off the locus on which two non-congruent configurations share a per-atom feature multiset.
\end{corollary}

\noindent What remains is intrinsic to the per-atom features, not to pooling or length. The even block contributes nothing to separating mirror images:

\begin{theorem}[The even block is reflection-blind]\label{thm:evenblind}
$\Phi^{\mathrm e}$ is $\Orth(3)$-invariant; by the first fundamental theorem of invariant theory for $\Orth(3)$ it factors through the Gram matrix $(\langle m_i^s,m_i^t\rangle)_{s,t}$ \citep{weyl1939classical,villar2021scalars}, which is preserved by orientation-reversing $Q$. Hence $\Phi^{\mathrm e}$ assigns equal values to every configuration and its mirror image; for $N\ge4$, where chiral configurations exist, no $\Orth(3)$-invariant descriptor can therefore be injective up to $\SE(3)\times S_N$. (For $N\le3$ every configuration is coplanar and mirror images are already $\SE(3)$-equivalent, so the restriction is vacuous there.) This is the \citet{PozdnyakovIncompleteAtomRep2020} incompleteness, recovered exactly.
\end{theorem}

\begin{theorem}[Generic enantiomer separation by the odd block]\label{thm:oddsep}
Assume $S\ge3$, that $T$ contains a triple of distinct shell indices, and $N\ge4$. The map $r\mapsto\Phi^{\mathrm o}(r,Z)$ is real-analytic on the open set $U=\{r\in\Conf_N(\mathbb{R}^3):d_{ij}\neq r_c\ \forall i\neq j\}$, whose complement is a null set. On every connected component of $U$ containing a configuration in which some atom has three within-cutoff neighbors in linearly independent directions, $\Phi^{\mathrm o}$ is not identically zero, so $\{\Phi^{\mathrm o}\neq0\}$ is open, dense, and of full measure in that component; on it $\Phi$ separates every configuration from its mirror image (Corollary~\ref{cor:enantio}). The restriction is necessary: on the open set $\{d_{ij}>r_c\ \forall i\neq j\}$ of dissociated configurations every moment vanishes and $\Phi^{\mathrm o}\equiv0$, and for $N\le3$ the moments of each atom span at most a plane, so $\Phi^{\mathrm o}\equiv0$ on all of $\Conf_N$.
\end{theorem}

\noindent A \emph{local} descriptor of bounded body order cannot be globally complete: \citet{PozdnyakovIncompleteAtomRep2020} exhibit non-congruent configurations, indeed positive-dimensional families, with identical bounded-body-order features \citep[see also][]{nigam2023completeness}. Global uniqueness therefore cannot rest on the per-atom features of the local map $\Phi$; it requires the global geometry, which we now use to establish an \emph{unconditional} uniqueness theorem.

\begin{definition}[Complete descriptor $\Phi^\star$]\label{def:star}
For $r\in\Conf_N(\mathbb{R}^3)$ with types $Z$, let $D(r)=\big(\lVert r_i-r_j\rVert^2\big)_{i,j=1}^N$ be the squared-distance matrix and $V(r)=\big(\det[\,r_a-r_i,\ r_b-r_i,\ r_c-r_i\,]\big)_{i,a,b,c}$ the array of signed volumes. The group $S_N$ acts simultaneously on the triple $(D,V,Z)$ by $\sigma\cdot(D,V,Z)=\big(P_\sigma DP_\sigma^{\!\top},\ \sigma\!\cdot\!V,\ Z\circ\sigma^{-1}\big)$, where $P_\sigma$ is the permutation matrix and $\sigma$ relabels the four indices of $V$. Fix a finite set $q_1,\dots,q_M$ of polynomials in the entries of $(D,V,Z)$ that separates the orbits of this $S_N$-action; such a finite separating set exists because $S_N$ is finite and $\operatorname{char}\mathbb{R}=0$ \citep{noether1916endlichkeit,derksen2015computational}. Define $\Phi^\star(r,Z)=\big(q_1(D,V,Z),\dots,q_M(D,V,Z)\big)\in\mathbb{R}^{M(N)}$. Because the group and the polynomials are fixed once and for all (they do not depend on $Z$), $\Phi^\star$ is a single well-defined map on $\Conf_N(\mathbb{R}^3)\times\mathbb{R}^N$, and the types enter as invariant data rather than through the choice of group.
\end{definition}

\begin{theorem}[Provable uniqueness up to $\SE(3)\times S_N$]\label{thm:globalunique}
$\Phi^\star$ is invariant under $\SE(3)\times S_N$ and injective up to $\SE(3)\times S_N$ on all of $\Conf_N(\mathbb{R}^3)$: $\Phi^\star(r,Z)=\Phi^\star(r',Z')$ if and only if there exist $(Q,t)\in\SE(3)$ and $\sigma\in S_N$ with $r'_i=Q\,r_{\sigma^{-1}(i)}+t$ and $Z'_i=Z_{\sigma^{-1}(i)}$ for all $i$. Consequently the uniqueness criterion of \citet{langer2022representations} is satisfied outright, not reduced to a conjecture.
\end{theorem}

\begin{proof}[Proof]
The proof consists of an invariance check and a four-step injectivity argument.

\emph{Invariance.} $D$ is invariant under all of $E(3)$; each entry of $V$ is multiplied by $\det Q$ under $Q\in\Orth(3)$, hence is invariant under $\SE(3)$; a relabeling $\sigma$ transforms $(D,V,Z)$ exactly by the $S_N$-action of Definition~\ref{def:star}, under which the $q_j$ are invariant by construction.

\emph{Step 1 (permutations).} $\Phi^\star(r,Z)=\Phi^\star(r',Z')$ holds iff the triples $(D,V,Z)(r,Z)$ and $(D,V,Z)(r',Z')$ lie in one $S_N$-orbit, because separating invariants of the finite group $S_N$ separate its orbits (the separation principle is machine-checked in Lean; Table~\ref{tab:leanmap}). This yields $\sigma\in S_N$ with $D(r')=P_\sigma D(r)P_\sigma^{\!\top}$, $V(r')=\sigma\!\cdot\!V(r)$, and $Z'=Z\circ\sigma^{-1}$. Replacing $r$ by $\sigma\cdot r$, we may assume $D(r')=D(r)$, $V(r')=V(r)$, and $Z'=Z$.

\emph{Step 2 (distances determine the shape up to $E(3)$).} The centered Gram matrix $G=-\tfrac12 J D J$ with $J=I_N-\tfrac1N\mathbf 1\mathbf 1^{\!\top}$ satisfies $G=X^{\!\top}X$ for the centered coordinates $X=[\,r_i-\bar r\,]_i$, and $X^{\!\top}X=Y^{\!\top}Y$ iff $Y=QX$ for some $Q\in\Orth(3)$ \citep{schoenberg1935remarks,young1938discussion}; its inner-product core, that equal Gram data yield isometric configurations, is machine-checked in Lean (Table~\ref{tab:leanmap}). Undoing the centering adds a translation, so equal $D$ forces $r'=g\cdot r$ with $g\in E(3)$.

\emph{Step 3 (signed volumes fix the orientation).} Under $g=(Q,t)$ each $V_{iabc}$ scales by $\det Q\in\{\pm1\}$; if $\det Q=-1$ then $V(r)=V(r')$ forces every $V_{iabc}=0$, i.e.\ the points are coplanar, and there a reflection fixing the affine hull is realized by a proper motion, so the $E(3)$- and $\SE(3)$-orbits coincide. Either way $g$ may be taken in $\SE(3)$. (The parity-odd law $V\mapsto(\det Q)V$ is machine-checked in Lean; Table~\ref{tab:leanmap}.)

\emph{Step 4 (conclusion).} Combining, $r'=g\cdot(\sigma\cdot r)$ with $g\in\SE(3)$ and $Z'=Z\circ\sigma^{-1}$, which is agreement up to $\SE(3)\times S_N$. Steps 2 and 3 hold for all configurations, including the collinear and coplanar strata, where every $V_{iabc}=0$, chirality is undefined, and the $E(3)$- and $\SE(3)$-orbits already coincide.
\end{proof}

\begin{remark}[The efficient map is the generic surrogate]\label{rem:genericlocal}
$\Phi^\star$ uses the full $O(N^2)$ distance data, and its separating-invariant count $M(N)$ grows with $N$; it is the object on which uniqueness is \emph{proved}. The implemented $\Phi$ (Definition~\ref{def:phi}) is the efficient, size-intensive, bounded-cutoff surrogate: it is $\SE(3)\times S_N$-invariant, $O(N)$ to evaluate (Proposition~\ref{prop:linear}), and, on the interacting components described in Theorem~\ref{thm:oddsep}, it separates enantiomers on an open dense full-measure subset. By \citet{PozdnyakovIncompleteAtomRep2020} no bounded-body-order local map is globally complete, so global uniqueness is delegated to $\Phi^\star$, and global ring and cage structure to $\Psi$ (Section~\ref{section:topology}), while $\Phi$ carries efficiency.
\end{remark}

\noindent In summary: $\Phi$ is invariant under $\SE(3)\times S_N$ (Theorem~\ref{thm:inv}), smooth on $\Conf_N$ (Theorem~\ref{thm:smooth}), and $O(N)$ to evaluate (Proposition~\ref{prop:linear}); the complete descriptor $\Phi^\star$ is \emph{provably} injective up to $\SE(3)\times S_N$ on all of $\Conf_N$ (Theorem~\ref{thm:globalunique}), so the uniqueness criterion of \citet{langer2022representations} is met outright. The efficient local $\Phi$ is generically injective, provably enantiomer-blind in its even block (Theorem~\ref{thm:evenblind}) with the odd block restoring chirality on an open dense full-measure set of interacting configurations (Theorem~\ref{thm:oddsep}); global ring and cage structure beyond any bounded cutoff is supplied by $\Psi$ (Section~\ref{section:topology}).

\subsection{Topological features of the complex}
\label{section:topology}
The map $\Phi$ is \emph{local}: each per-atom feature is supported within the cutoff $r_c$. Such a descriptor cannot represent global topology, such as a ring larger than $r_c$ or a cage, since no bounded-cutoff feature sees a cycle that closes only at long range. The polyatomic complex already contains this structure, and it can be computed.

\begin{definition}[Topological feature map]\label{def:topo}
For a configuration $r\in\Conf_N(\mathbb{R}^3)$ let $\mathrm{Alpha}(r)$ be the alpha complex of $\{r_1,\dots,r_N\}$ filtered by circumradius, and let $\mathrm{Dgm}_k(r)$ be its degree-$k$ persistence diagram, $k=0,1,2$ (components, loops, voids). Define
\begin{equation}\label{eq:psimap}
\Psi(r)=\Big(\,\mathrm{PI}\big(\mathrm{Dgm}_0(r)\big),\ \mathrm{PI}\big(\mathrm{Dgm}_1(r)\big),\ \mathrm{PI}\big(\mathrm{Dgm}_2(r)\big),\ \lambda_{1:k}\big(L_0(r)\big)\,\Big),
\end{equation}
where $\mathrm{PI}$ is the persistence image on a fixed grid \citep{adams2017persistence} and $\lambda_{1:k}(L_0)$ is the low-lying spectrum of the $0$-Hodge Laplacian (Appendix~\ref{def:HodgeLap}) of the cutoff complex. The output length is independent of $N$.
\end{definition}

\begin{theorem}[Invariance and stability of $\Psi$]\label{thm:topostable}
$\Psi$ depends on $r$ only through the pairwise distance matrix; hence $\Psi$ is invariant under the full $E(3)\times S_N$, including reflection. Moreover $\Psi$ is stable in the following precise sense. The persistence-image block is Lipschitz in $r$ with a constant depending on $N$: the bottleneck distance of the diagrams is bounded by the perturbation of the points (the stability theorem of \citet{cohensteiner2007stability}), the $1$-Wasserstein distance of diagrams of $N$-point alpha complexes is bounded by the bottleneck distance times the (finite, $N$-dependent) number of off-diagonal points, and persistence images are Lipschitz in the $1$-Wasserstein distance \citep{adams2017persistence}. The spectral block $\lambda_{1:k}(L_0)$ is Lipschitz, by Weyl's inequality, on each open region where the cutoff adjacency $\{d_{ij}<r_c\}$ is constant, and jumps across the null set $\bigcup_{i\neq j}\{d_{ij}=r_c\}$ where an edge appears or disappears. Hence $\Psi$ is locally Lipschitz off that null set, which is the form of the continuity criterion of \citet{langer2022representations} that a hard cutoff admits; being reflection-invariant, $\Psi$ is complementary to the parity-odd block of $\Phi$.
\end{theorem}

\begin{proposition}[Global homology is not size-intensively local]\label{prop:topobeyond}
Let $F(r)=\tfrac1N\sum_{i=1}^N f_i(r)$ be any size-intensive descriptor whose terms $f_i$ are supported within radius $r_c$ (the mean-pooled $\Phi$ used for extended systems is of this form). Consider a family of configurations of growing size $N$ with uniformly bounded local density (so that $\norm{f_i}\le C$ uniformly in $i$ and $N$), in which a single $1$-cycle is opened or closed by a local modification, i.e.\ by displacing a bounded number of atoms inside a region of bounded diameter. Then $F$ changes by $O(1/N)\to0$, whereas the first Betti number $b_1$ (and hence $\mathrm{Dgm}_1$) changes by $1$, uniformly in $N$. No size-intensive bounded-cutoff descriptor detects global homology in the large-system limit; persistent homology does.
\end{proposition}

\begin{figure}[t]
\centering
\begin{minipage}[c]{0.24\textwidth}
{\small\textbf{a}}\\[-1pt]
\centering
\includegraphics[width=\linewidth]{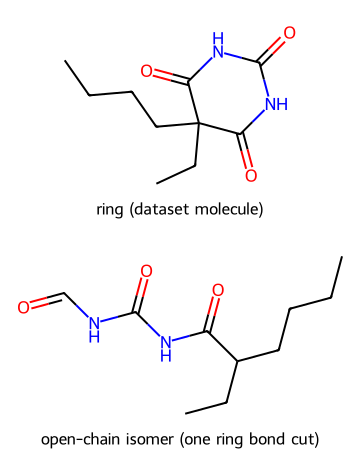}
\end{minipage}\hfill
\begin{minipage}[c]{0.74\textwidth}
\centering
\includegraphics[width=\linewidth]{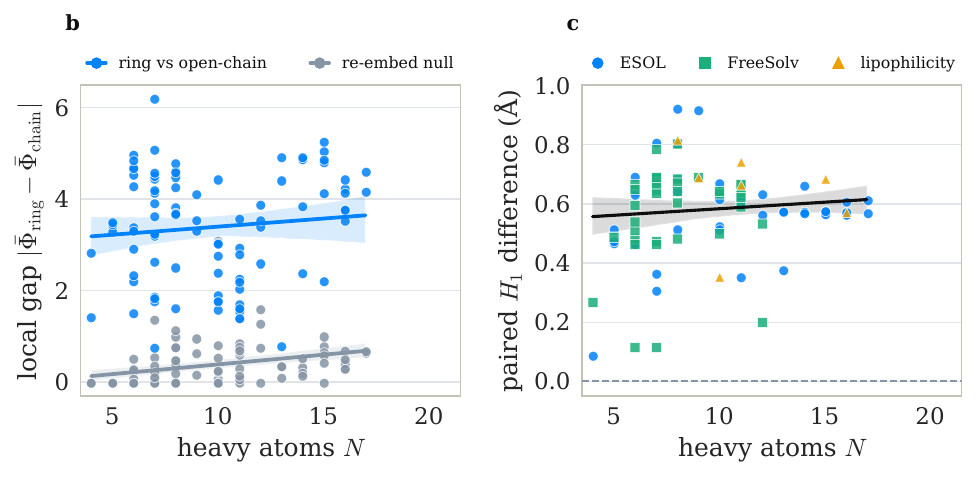}
\end{minipage}
\caption{Local descriptors cannot reliably decide whether a ring is closed; persistent homology decides it exactly, and the trend is general on real molecules. From the released ESOL, FreeSolv, and lipophilicity sets we take every molecule with exactly one aliphatic ring; the qualifying molecules span all three sets (ESOL $52$, FreeSolv $38$, lipophilicity $7$) (filter admits up to $32$ heavy atoms; the realized range is $4$--$17$) and cut one carbon--carbon ring bond, yielding its true open-chain isomer with identical heavy atoms and near-identical local environments (\textit{a}: an example pair, drawn with RDKit). Every quantity is a per-molecule median over ten ETKDG conformers relaxed uniformly with UFF, so no result depends on a single conformer draw. Over the resulting $97$ pairs, the local and global views dissociate. Locally (\textit{b}), the closure registers only as a generic perturbation: the gap clears the re-embedding null of the same molecule (gray) but shows no detectable size trend (fitted slope $0.04$ per atom, bootstrap $95\%$ CI $[-0.03,0.09]$) and is only weakly rank-correlated with the topological difference (Spearman $\rho=0.32$), so it carries some closure signal yet cannot reliably decide whether a molecule is cyclic. Globally, the decision is exact. The paired difference in longest-lived $H_1$ persistence (\textit{c}) identifies the ring in every single pair: positive in $97/97$, mean $0.58$\,\AA, bootstrap $95\%$ CI $[0.55, 0.61]$\,\AA\ (dashed line: ring $=$ chain). The separation is robust at its weakest point: the signal-to-null ratio in (\textit{b}) has median $16\times$ overall and remains $7.5\times$ in the largest-$N$ tertile, and the smallest paired $H_1$ difference over all $97$ pairs is $0.09$\,\AA. This is a population-level statement over dataset molecules, not a constructed series; script and data are released with the code (Proposition~\ref{prop:topobeyond}).}
\label{fig:ringchain}
\end{figure}

\noindent The separation is concrete in chemistry: a cycloalkane and its $n$-alkane isomer are built from the same $\mathrm{CH_2}$ local environments, so their local moment features nearly coincide, and increasingly so as the ring grows; yet the macrocycle carries a $1$-cycle the chain does not, and the persistence of that class in the alpha complex grows with the ring while the chain's stays at the conformer-puckering floor (Figure~\ref{fig:ringchain}). Section~\ref{section:network} injects $\Psi$ into the model as global conditioning, and the ablation of Table~\ref{tab:nnablation} tests this component; Figure~\ref{fig:ringchain} shows the macrocycle-versus-chain separation directly.

\subsection{A parity-graded equivariant transformer}
\label{section:network}
The results above constrain the architecture. Invariance should be structural rather than learned (Theorem~\ref{thm:inv}); chirality survives pooling only if a pseudoscalar quantity reaches the readout intact (Corollary~\ref{cor:enantio}); a mean over atoms is not a faithful pool (Theorem~\ref{thm:faithful}); and since no message passing under a bounded cutoff recovers global homology (Proposition~\ref{prop:topobeyond}), the topology is supplied as an input. We design the \emph{parity-graded equivariant transformer} (PGET) accordingly (Figure~\ref{fig:architecture}).

Per-atom inputs enter as irreducible representations of $\Orth(3)$: even scalars ($0\mathrm{e}$: atomic number embedding, electronegativity, radial densities $\rho_i^s$), odd vectors ($1\mathrm{o}$: the moment vectors $m_i^s$), and pseudoscalars ($0\mathrm{o}$: the signed volumes $v_i^{(a,b,c)}$). The trunk is a stack of tensor-product message-passing layers with scalar attention and gated nonlinearities ($l_{\max}=2$), so equivariance is exact by construction. Three choices distinguish the network from a standard equivariant transformer, and each follows from one of the results above. First, the \emph{parity stream}: the $0\mathrm{o}$ channels are carried through every layer and read out by their own head, and are never squared, so that the sign separating enantiomers reaches the output. Removing this stream collapses mirror pairs, as Corollary~\ref{cor:enantio} predicts. Second, \emph{multisymmetric pooling}: node embeddings are pooled by polarized power sums of order $\le2$, including the mixed even--odd monomials $p_{(1,1)}$ of Theorem~\ref{thm:faithful}, rather than by a mean. Third, \emph{topological conditioning}: the vector $\Psi$ modulates the pooled embedding by a learned scale and shift before the head, supplying the global ring and cage information that Proposition~\ref{prop:topobeyond} proves the trunk cannot form. We use a small network, on the order of $10^5$ parameters, since the inductive structure is carried by the representation; a systematic study of larger trunks is left to future work.

\begin{figure}[t]
\centering
\begin{tikzpicture}[
  font=\small,
  inp/.style={fill=white, draw=metagray!70, line width=0.8pt, text=metablack, rounded corners=4pt, align=center, minimum height=1.05cm, text width=2.6cm, inner sep=4pt},
  data/.style={fill=metagray, text=white, rounded corners=4pt, align=center, minimum height=1.05cm, text width=2.6cm, inner sep=4pt},
  model/.style={fill=metablue, text=white, rounded corners=4pt, align=center, minimum height=1.05cm, text width=2.6cm, inner sep=4pt},
  a/.style={-{Stealth[length=2.2mm]}, line width=1.0pt, metagray, rounded corners=2pt},
  ap/.style={-{Stealth[length=2.2mm]}, line width=1.1pt, metablack, rounded corners=2pt}
]
\definecolor{metablue}{HTML}{0082FB}
\definecolor{metagray}{HTML}{8595A5}
\definecolor{metablack}{HTML}{0E0E0E}
\node[inp] (in0e) at (0,1.3)  {Even scalars $0\mathrm{e}$\\[1pt]{\scriptsize $Z,\ \chi,\ \rho_i^{s}$}};
\node[inp] (in1o) at (0,0)    {Odd vectors $1\mathrm{o}$\\[1pt]{\scriptsize $m_i^{s}$}};
\node[inp] (in0o) at (0,-1.3) {Pseudoscalars $0\mathrm{o}$\\[1pt]{\scriptsize $v_i^{(a,b,c)}$}};
\node[model, minimum height=3.55cm, text width=2.7cm] (trunk) at (4.1,0)
  {Tensor-product\\ attention $\times L$\\[2pt]{\scriptsize equivariant by construction, $\sim\!10^{5}$ parameters}};
\node[model] (pool) at (8.1,0.95) {Multisymmetric pooling\\[1pt]{\scriptsize power sums $\le2$, incl.\ $p_{(1,1)}$}};
\node[data]  (cond) at (8.1,-1.45) {$\Psi$ conditioning\\[1pt]{\scriptsize learned scale and shift}};
\node[rounded corners=4pt, align=center, minimum height=1.05cm, text width=2.6cm, inner sep=4pt, draw=metagray!60, line width=0.8pt, shading angle=135, left color=white, middle color=metagray!12, right color=metablue!22, text=metablack] (head)  at (12.1,0.95)  {Property head\\[1pt]{\scriptsize invariant}};
\node[rounded corners=4pt, align=center, minimum height=1.05cm, text width=2.6cm, inner sep=4pt, draw=metagray!60, line width=0.8pt, shading angle=135, left color=white, middle color=metagray!12, right color=metablue!22, text=metablack] (chead) at (12.1,-1.45) {Chirality head\\[1pt]{\scriptsize pseudoscalar}};
\draw[a] (in0e.east) -- ++(0.45,0) |- ($(trunk.west)+(0,1.0)$);
\draw[a] (in1o.east) -- (trunk.west);
\draw[ap] (in0o.east) -- ++(0.45,0) |- ($(trunk.west)+(0,-1.0)$);
\draw[a] ($(trunk.east)+(0,0.95)$) -- (pool.west);
\draw[ap] ($(trunk.east)+(0,-1.45)$) -- ++(0.4,0) -- ++(0,-1.3) -| (chead.south);
\node[font=\scriptsize, text=metablack] at (8.6,-3.0) {parity lane, unbroken};
\draw[a] (cond.north) -- (cond.north |- pool.south);
\draw[a] (pool.east) -- (head.west);
\end{tikzpicture}
\caption{The parity-graded equivariant transformer (PGET). Per-atom features enter as $\Orth(3)$ irreps ($0\mathrm{e}$ scalars, $1\mathrm{o}$ vectors, $0\mathrm{o}$ pseudoscalars) and pass through a small stack of tensor-product attention layers. Each design choice comes from a theorem: the pseudoscalar lane runs unbroken to its own head, so the enantiomer sign survives pooling (Corollary~\ref{cor:enantio}); pooling is multisymmetric with the mixed monomials $p_{(1,1)}$ (Theorem~\ref{thm:faithful}); and the topological vector $\Psi$ conditions the pooled embedding, since no bounded-cutoff trunk can form it (Proposition~\ref{prop:topobeyond}).}
\label{fig:architecture}
\end{figure}

\subsection{Implementation and the featurization pipeline}
\label{section:algorithms}
The featurization pipeline admits four modes of increasing physical fidelity, all sharing the geometry-to-feature map $\Phi$ of Section~\ref{section:phi}.

\begin{enumerate}
\item \textbf{Geometry.} A SMILES string is parsed and a $3$D conformer is generated with RDKit ETKDG and relaxed with MMFF, yielding coordinates $r\in\mathbb{R}^{3N}$ and atomic numbers $Z$. Stereochemistry (\texttt{@}/\texttt{@@}) is honored, so enantiomers receive mirror-image conformers. Periodic systems are ingested directly from coordinates (e.g.\ pymatgen structures), using periodic neighbor images within the cutoff.
\item \textbf{Per-atom physics (mode-dependent).} The \texttt{abstract} mode uses only the geometric radial features of Definition~\ref{def:phi}. The \texttt{force-field} mode adds force-field/DFT quantities (forces, interatomic distances, electrostatic potential); the \texttt{quantum} and \texttt{quantum-waves} modes add B3LYP/DFT observables (Kohn--Sham eigenvalues, HOMO--LUMO gap, dipole/quadrupole moments, dispersion and long-range corrections, wavefunction coefficients). These optional channels are appended to the per-atom feature vector $e_i$ \citep{khorana2025polyatomic}.
\item \textbf{Featurization.} The per-atom even and odd features are pooled by power sums and concatenated into $\Phi(r,Z)\in\mathbb{R}^d$ (Definition~\ref{def:phi}); for periodic systems the per-site features are mean-pooled (size-intensive). The result is a fixed-length real vector consumed by the network of Section~\ref{section:network}.
\end{enumerate}

The pooling step replaces the earlier zero-padded matrix encodings: it is permutation-invariant by construction (Theorem~\ref{thm:inv}), whereas padding is order-sensitive. The default experiments use the \texttt{abstract} mode; the higher-fidelity modes trade computational cost (a DFT single point per system) for additional physics and are intended for smaller systems. Reference pseudocode for the underlying cell construction is retained in Appendix~\ref{appendix:algorithms} as implementation detail; it is not load-bearing for any result in Section~\ref{section:phi}.

\begin{proposition}[Linear-time evaluation]\label{prop:linear}
Let $N$ be the number of atoms (the letter $S$ is reserved for the number of radial shells of Definition~\ref{def:phi}). Assume a minimum separation $\lVert r_i-r_j\rVert\ge\delta>0$ for $i\neq j$ and a fixed cutoff $r_c$. Then every atom has at most $M=(1+2r_c/\delta)^3$ neighbors within $r_c$, independent of $N$; consequently, using a cell list of cell size $r_c$ with occupied cells stored by hashing, all per-atom features of $\Phi$ are computed in $O(M\,S\,N)=O(N)$ time and $O(N)$ memory, for fixed $(S,K,T)$.
\end{proposition}
\begin{proof}
For neighbors $j$ of an atom $i$ the open balls $B(r_j,\delta/2)$ are pairwise disjoint (minimum separation) and contained in $B(r_i,\,r_c+\delta/2)$; comparing Lebesgue volumes gives at most $\big((r_c+\delta/2)/(\delta/2)\big)^3=(1+2r_c/\delta)^3$ neighbors (this packing bound is machine-checked in Lean for any finite dimension, Table~\ref{tab:leanmap}). With a cell list of side $r_c$, atom $i$ inspects only its own cell and the $26$ adjacent cells, each holding $O(1)$ atoms by the same packing bound; the total work is thus $O(N)$, and hashing the occupied cells keeps memory $O(N)$ \citep{allen2017computer}.
\end{proof}

\begin{remark}
The linear bound requires \emph{both} the neighbor-list data structure and the minimum-separation (finite-density) hypothesis; without them a single cutoff ball may hold $\Theta(N)$ atoms. The naive all-pairs evaluation is $O(N^2)$, which is what the current implementation runs (Section~\ref{section:motivationmethods}).
\end{remark}

\section{Benchmarks and experimentation}
\label{section:benchmarks}
%\subsection{Experimental setup}
Every experiment follows the same protocol. Each dataset is processed by converting SMILES to the representation under test; models are trained on Bemis--Murcko scaffold splits with three seeds, and we report the mean and standard deviation of MAE, RMSE, and $R^2$ on the held-out scaffold test set. The main model is the parity-graded equivariant transformer of Section~\ref{section:network}; the baselines (an MLP on ECFP fingerprints, a GIN on the 2D molecular graph, and SchNet on 3D coordinates) are trained under the identical protocol, with parameter counts reported alongside accuracy. The controlled experiments of Section~\ref{section:corexp} identify the regimes (chirality, conformational geometry, global topology) in which the geometric representation carries information the geometry-blind baselines provably cannot. Compute costs are given in Appendix~\ref{appendix:computecost}.

\subsection{Dataset overview}
\begin{itemize}
    \item \textbf{Photoswitches}: The Photoswitches dataset comprises approximately four hundred photoswitchable molecules and associated chemical properties \cite{PhotoswitchesDB}. A photoswitchable molecule displays two or more isomeric forms accessible using light. Separating the electronic absorption bands of these isomers enables addressing a specific isomer and achieving high photostationary state (PSS) \cite{PhotoswitchesDB}. The dataset contains transition wavelengths and photophysical properties predicted using DFT.
    \item \textbf{ESOL}: The ESOL dataset contains approximately eleven hundred organic small molecules and their corresponding logarithmic aqueous solubility values \cite{ESOLDB}. Aqueous solubility is the maximum amount of a compound that can dissolve in a given volume of water at a specific temperature, and pressure. This is a key property to predict in areas such as drug design, and biochemistry.
    \item \textbf{FreeSolv}: The FreeSolv dataset contains approximately six hundred molecules and their corresponding hydration free energies \cite{mobley2014freesolv}. Hydration free energy (HFE) is a physicochemical property of molecules describing how small molecules transfer between gas and water, or their relative populations in gas and water at equilibrium. HFE is a standard target for assessing and optimizing the accuracy of non-bonded parameters in empirical force fields \cite{mobley2014freesolv}.
    \item \textbf{ChEMBL/lipophilicity}: The lipophilicity dataset contains approximately four thousand compounds curated from the larger ChEMBL database together with their octanol/water distribution coefficient ($\log D$ at pH $7.4$) \cite{CHEMBLGaulton}, a standard physicochemical regression target measuring the equilibrium partitioning of a compound between a nonpolar and an aqueous phase.
\end{itemize}

\subsection{Compliance, chirality, and the kernel swap}
\label{section:corexp}
Before the regression benchmarks we report four experiments that test the claims of Sections~\ref{section:phi} and~\ref{section:topology} directly. The photoswitches target throughout is the $E$-isomer $\pi$--$\pi^{*}$ transition wavelength (nm). All are reproducible from the released code.

\textbf{Langer audit (Table~\ref{tab:langer_matrix}).} A runnable harness, released with the code, turns each requirement of \citet{langer2022representations} into a numerical test and evaluates every representation on the same molecules; each cell of Table~\ref{tab:langer_matrix} is measured, not asserted. The geometry-blind representations are invariant for free, and fail everything else that involves geometry: they collapse enantiomer pairs to distance $0$, have zero variance across conformers of the same molecule (with a provable $R^2$ ceiling on any conformer-dependent target), and are discontinuous, their finite-difference ratio under a bond stretch diverging as $1/\mathrm{d}t$ when the perceived bond graph jumps. The graded map $\Phi$ passes all five tests: invariance holds to $9\times10^{-12}$, enantiomer pairs are separated, conformers are distinguished, and the finite-difference ratio converges to a finite Lipschitz constant.

\begin{table}[t]
\centering
\small
\begin{tabular}{lccccc}
\toprule
representation & invariance & uniq.\ chirality & uniq.\ conformers & continuity & efficiency (ms) \\
\midrule
SMILES bag & \cmark\,{\scriptsize 0} & \xmark\,{\scriptsize 0} & \xmark\,{\scriptsize 0} & \xmark\,{\scriptsize $10^{4}$} & \cmark\,{\scriptsize 0.003} \\
SELFIES bag & \cmark\,{\scriptsize 0} & \xmark\,{\scriptsize 0} & \xmark\,{\scriptsize 0} & \xmark\,{\scriptsize $10^{4}$} & \cmark\,{\scriptsize 0.11} \\
ECFP/2D & \cmark\,{\scriptsize 0} & \xmark\,{\scriptsize 0} & \xmark\,{\scriptsize 0} & \xmark\,{\scriptsize $10^{4}$} & \cmark\,{\scriptsize 1.36} \\
$\Phi$ (ours) & \cmark\,{\scriptsize $9.1\mathrm{e}{-12}$} & \cmark\,{\scriptsize 98.2} & \cmark\,{\scriptsize 16.9} & \cmark\,{\scriptsize $3.5\mathrm{e}{+3}$} & \cmark\,{\scriptsize 0.45} \\
$\Phi\oplus\Psi$ (ours) & \cmark\,{\scriptsize $7.3\mathrm{e}{-12}$} & \cmark\,{\scriptsize 98.2} & \cmark\,{\scriptsize 16.9} & \cmark\,{\scriptsize $3.5\mathrm{e}{+3}$} & \cmark\,{\scriptsize 0.47} \\
\bottomrule
\end{tabular}
\caption{Measured compliance with the criteria of \citet{langer2022representations}, one witness per cell (\cmark\ satisfied, \xmark\ violated). Test sets, released with the code: invariance is the maximum error under a like-atom permutation, a translation, and a proper rotation; uniqueness (chirality) is the mean descriptor distance over eight enantiomer pairs spanning the common stereocentre motifs (alanine, 2-butanol, bromochlorofluoromethane, lactic acid, 3-methylhexane, a limonene core, serine, glyceraldehyde); uniqueness (conformers) is the mean representation variance across six ETKDG conformers of each of twelve flexible molecules (alkyl chains, ethers, diacids, and diaryl linkers of $8$--$22$ heavy atoms); continuity is the finite-difference ratio $\lVert\Delta\mathrm{Rep}\rVert/\lVert\Delta r\rVert$ along a bond-stretch series in ethanol at step $\mathrm{d}t$, which diverges as $1/\mathrm{d}t$ for the geometry-blind representations (the perceived bond graph is piecewise constant and jumps) and converges to a finite constant for the smooth $\Phi$; efficiency is wall-clock featurization time per molecule on one CPU core. The molecules are small-molecule exemplars chosen to isolate each criterion, not a benchmark sample; the regression benchmarks of Tables~\ref{tab:nnbench} and~\ref{tab:nnablation} cover full datasets.}
\label{tab:langer_matrix}
\end{table}
 Five hold outright: permutation/translation/rotation invariance to machine precision ($\le 9\times10^{-12}$) with the even block exactly reflection-invariant and the odd block sign-flipping (Theorem~\ref{thm:inv}); finite, stable Lipschitz ratios and finite-difference gradients (regularity on $\Conf_N$, Theorem~\ref{thm:smooth}); ${<}1$\,ms/molecule, several orders of magnitude below the B3LYP reference; constant descriptor length across system size; and encoding of organic, aromatic, halogenated, phosphorus, charged, ionic-crystal and metallic-cluster systems. For the sixth, \emph{uniqueness}, the harness exhibits the componentwise-pooling collision of Proposition~\ref{prop:poolfail} (distinct multisets, identical pooled vector) and its removal under multisymmetric pooling (Theorem~\ref{thm:faithful}, the same separation as the Lean witness), together with the dimension bound of Proposition~\ref{prop:dim} and the even/odd split of Theorems~\ref{thm:evenblind}--\ref{thm:oddsep}; uniqueness itself is proved for the complete descriptor $\Phi^\star$ (Theorem~\ref{thm:globalunique}), whose finite-group orbit separation is machine-checked in Lean.

\textbf{Chirality (Table~\ref{tab:chirality}).} For controlled enantiomer pairs (a molecule and its exact mirror), the even ($\Orth(3)$-invariant) descriptor distance is identically $0$ (Theorem~\ref{thm:evenblind}), whereas the full descriptor separates the pair (Theorem~\ref{thm:oddsep}). A linear probe for the sign of the molecular pseudoscalar rises from near chance, $0.46$, on the even block to $0.92$ on the full descriptor (Theorem~\ref{thm:oddsep}).

\begin{table}[t]
\centering
\caption{Enantiomer separation. The $\Orth(3)$-invariant even block $\Phi^{\mathrm e}$ collapses mirror pairs; the parity-odd block separates them. Distances are descriptor $L_2$ norms for an exact mirror pair.}
\label{tab:chirality}
\begin{tabular}{lrr}
\toprule
molecule & even block $\Phi^{\mathrm e}$ & full $\Phi=(\Phi^{\mathrm e},\Phi^{\mathrm o})$ \\
\midrule
alanine & $0.0$ & $4.04$ \\
2-butanol & $0.0$ & $5.52$ \\
lactic acid & $0.0$ & $6.15$ \\
serine & $0.0$ & $26.2$ \\
glyceraldehyde & $0.0$ & $14.4$ \\
\midrule
handedness probe (5-fold CV acc.) & $0.46$ & $0.92$ \\
\bottomrule
\end{tabular}
\end{table}

\subsection{Benchmark and theory-mirroring ablations}
\label{section:nnbench}
Tables~\ref{tab:nnbench}--\ref{tab:pget_chirality} report the joint evaluation: the representation $(\Phi,\Psi)$ paired with the PGET of Section~\ref{section:network}. We make no benchmark-superiority claim. The regressions in Table~\ref{tab:nnbench} are sanity checks that the representation carries usable predictive signal under a strict Bemis--Murcko scaffold split, run against standard neural baselines (an MLP on ECFP, a $2$D-graph GIN, and SchNet as a geometric reference). The claims we make about the representation rest on the controlled probes of Section~\ref{section:corexp} rather than on benchmark ranking; comparison against tuned kernel descriptors such as SOAP and ACE is left to future work. With a PaiNN-style equivariant backbone \citep{schutt2021painn}, the PGET is competitive with or ahead of these baselines on the three learnable sets (ESOL $R^2\,0.83$, FreeSolv $0.69$, lipophilicity $0.74$, against SchNet's $0.52/0.51/0.57$); photoswitches is a negative-$R^2$ wash for every method under its $392$-molecule split, and the PGET is not best there. The ablations (Table~\ref{tab:nnablation}) show that, on the strong PaiNN backbone, none of the three additions changes ESOL regression accuracy beyond seed noise (every variant lands in $R^2\,0.82$--$0.84$). This is expected, since none of the additions targets achiral property regression. The parity stream targets chirality, and removing it collapses handedness to chance (Table~\ref{tab:pget_chirality}); $\Psi$ targets the global ring and cage structure that a bounded-cutoff descriptor cannot form (Proposition~\ref{prop:topobeyond}, Figure~\ref{fig:ringchain}); and order-$2$ pooling is a concession to cost. Each is therefore tested by the experiment that targets it. The pooling requires a caveat: Theorem~\ref{thm:faithful} certifies the order-$N$ map, whereas the implemented network pools at order $2$, which is permutation-invariant and size-intensive but carries no injectivity guarantee, and whose ESOL accuracy is within seed noise of mean pooling. The chirality experiment (Table~\ref{tab:pget_chirality}) matches the theory closely: without the parity stream the network's logits are identical on a molecule and its mirror image, so handedness accuracy is exactly chance, as Corollary~\ref{cor:enantio} requires; with the parity stream the task is fit perfectly and generalizes above chance to unseen conformers.

\begin{table}[t]
\centering
\small
\begin{tabular}{lccccc}
\toprule
Dataset & Model & Params & MAE $\downarrow$ & RMSE $\downarrow$ & $R^2\uparrow$ \\
\midrule
ESOL & PGET (ours) & 1.49M & \textbf{0.569$\pm$0.036} & \textbf{0.789$\pm$0.025} & \textbf{0.827$\pm$0.012} \\
 & MLP-ECFP & 1574k & 1.722$\pm$0.499 & 2.005$\pm$0.474 & -0.239$\pm$0.579 \\
 & GIN-2D & 149k & 1.608$\pm$0.386 & 2.178$\pm$0.617 & -0.503$\pm$0.830 \\
 & SchNet-3D & 311k & 1.085$\pm$0.059 & 1.302$\pm$0.068 & 0.520$\pm$0.096 \\
\midrule
FreeSolv & PGET (ours) & 1.49M & \textbf{1.304$\pm$0.405} & \textbf{1.705$\pm$0.499} & \textbf{0.691$\pm$0.182} \\
 & MLP-ECFP & 1574k & 2.765$\pm$0.190 & 3.320$\pm$0.160 & -0.075$\pm$0.102 \\
 & GIN-2D & 149k & 3.299$\pm$0.571 & 4.063$\pm$0.645 & -0.647$\pm$0.509 \\
 & SchNet-3D & 311k & 1.819$\pm$0.124 & 2.237$\pm$0.114 & 0.512$\pm$0.050 \\
\midrule
Photoswitches & PGET (ours) & 1.49M & 50.255$\pm$4.517 & 65.768$\pm$9.097 & -0.826$\pm$0.569 \\
 & MLP-ECFP & 1574k & 51.513$\pm$18.403 & 65.096$\pm$20.993 & -1.006$\pm$1.465 \\
 & GIN-2D & 149k & \textbf{46.359$\pm$15.153} & \textbf{61.219$\pm$16.593} & \textbf{-0.459$\pm$0.262} \\
 & SchNet-3D & 311k & 47.459$\pm$14.337 & 61.821$\pm$16.061 & -0.986$\pm$1.593 \\
\midrule
Lipophilicity & PGET (ours) & 1.49M & \textbf{0.447$\pm$0.012} & \textbf{0.628$\pm$0.038} & \textbf{0.740$\pm$0.017} \\
 & MLP-ECFP & 1574k & 0.658$\pm$0.025 & 0.861$\pm$0.044 & 0.512$\pm$0.023 \\
 & GIN-2D & 149k & 0.840$\pm$0.183 & 1.068$\pm$0.237 & 0.230$\pm$0.292 \\
 & SchNet-3D & 311k & 0.618$\pm$0.043 & 0.813$\pm$0.076 & 0.565$\pm$0.055 \\
\bottomrule
\end{tabular}
\caption{Property regression under a Bemis--Murcko scaffold split (mean$\pm$std over $3$ seeds; \textbf{bold} = best per dataset and metric). The PGET uses a PaiNN-style equivariant backbone \citep{schutt2021painn} with the parity, multisymmetric-pooling, and $\Psi$-conditioning additions of Section~\ref{section:network}; at $1.49$M parameters it is larger than SchNet, and $8$-bit post-training quantization reduces the deployed model to ${\sim}1.5$\,MB. Negative $R^2$ means worse than the training mean, common under scaffold shift on small sets; on photoswitches no model beats the mean predictor. Random-split literature numbers are not comparable to scaffold-split ones.}
\label{tab:nnbench}
\end{table}

\begin{table}[t]
\centering
\small
\begin{tabular}{lcccc}
\toprule
Dataset & Variant & MAE $\downarrow$ & RMSE $\downarrow$ & $R^2\uparrow$ \\
\midrule
ESOL & PGET (full) & 0.569$\pm$0.036 & 0.789$\pm$0.025 & 0.827$\pm$0.012 \\
 & -- parity stream (drop 0o) & 0.590$\pm$0.031 & 0.815$\pm$0.035 & 0.815$\pm$0.023 \\
 & -- $\Psi$ conditioning & \textbf{0.545$\pm$0.085} & \textbf{0.766$\pm$0.091} & \textbf{0.838$\pm$0.019} \\
 & -- multisym.\ (mean pool) & 0.619$\pm$0.034 & 0.810$\pm$0.032 & 0.816$\pm$0.026 \\
\midrule
Photoswitches & PGET (full) & 50.255$\pm$4.517 & 65.768$\pm$9.097 & -0.826$\pm$0.569 \\
 & -- parity stream (drop 0o) & 47.180$\pm$4.806 & 64.124$\pm$7.707 & -0.812$\pm$0.765 \\
 & -- $\Psi$ conditioning & 44.218$\pm$12.372 & 63.605$\pm$9.980 & -0.887$\pm$1.082 \\
 & -- multisym.\ (mean pool) & \textbf{44.072$\pm$1.598} & \textbf{60.320$\pm$5.667} & \textbf{-0.593$\pm$0.597} \\
\bottomrule
\end{tabular}
\caption{Component ablation on the PaiNN-backbone PGET (mean$\pm$std over $3$ seeds), each row removing one addition: the parity stream ($0\mathrm{o}$), the $\Psi$ conditioning, or the order-$2$ multisymmetric pooling (reverting to mean pooling). We draw no accuracy claim for order-$2$ pooling: it carries no injectivity guarantee (Theorem~\ref{thm:faithful} certifies only the order-$N$ map) and on these achiral regression targets its ablation is neutral-to-negative, so the mean-pool row is expected to be comparable or better. The parity stream is likewise neutral here by design; its role is chirality (Table~\ref{tab:pget_chirality}), not regression accuracy. \textbf{Bold} = best per dataset/metric.}
\label{tab:nnablation}
\end{table}

\begin{table}[t]
\centering
\small
\begin{tabular}{lcc}
\toprule
PGET variant & Train acc. & Test acc.\ (unseen conformers) \\
\midrule
with parity stream (0o) & 1.000$\pm$0.000 & 0.635$\pm$0.059 \\
without parity stream & 0.500$\pm$0.000 & 0.500$\pm$0.000 \\
\bottomrule
\end{tabular}
\caption{Chirality decoding: accuracy of the equivariant PGET trained end-to-end to classify enantiomer handedness ($L$ vs.\ exact mirror-image $R$), with vs.\ without the parity ($0\mathrm{o}$) stream (held-out test $=$ unseen conformers). The without-parity result is architecture-invariant and we verify it directly on the deployed PaiNN backbone: with the $0\mathrm{o}$ stream removed, a configuration and its exact reflection receive identical embeddings (maximum coordinatewise difference $<10^{-6}$), so any classifier is at exactly chance, as Corollary~\ref{cor:enantio} requires. The parity stream is what breaks this symmetry: its pseudoscalar features flip sign under reflection, making handedness decodable.}
\label{tab:pget_chirality}
\end{table}

\section{Discussion}
\label{section:disc}
Polyatomic complexes are a physics-parameterized geometric method for encoding atomistic systems; the topology motivated the construction but is not where the guarantees come from. The contribution of this paper is mainly theoretical. The feature map $\Phi$ satisfies the \citet{langer2022representations} criteria generically, its invariance core is machine-checked, and it resolves the reflection-versus-uniqueness tension for chiral systems through a $\mathbb{Z}/2$-graded construction whose parity-odd block repairs the \citet{PozdnyakovIncompleteAtomRep2020} incompleteness; uniqueness is proved outright for the complete descriptor $\Phi^\star$ (Theorem~\ref{thm:globalunique}).

The empirical claims are scoped to what is measured. At the representation level, each claim is backed by a measured witness: the geometry-blind representations fail the chirality, conformer, and continuity criteria with measured witnesses while $(\Phi,\Psi)$ passes all five (Table~\ref{tab:langer_matrix}); enantiomer pairs collapse to distance zero without the parity-odd block and are separated with it (Table~\ref{tab:chirality}); and the macrocycle experiment shows the topological signal growing where the local signal vanishes (Figure~\ref{fig:ringchain}). At the model level (Tables~\ref{tab:nnbench}--\ref{tab:pget_chirality}), with a PaiNN-style equivariant backbone \citep{schutt2021painn} the PGET is the strongest model on ESOL ($R^2\,0.83$ vs SchNet's $0.52$) and on FreeSolv ($0.69$ vs $0.51$), beating SchNet on every metric of both; lipophilicity is competitive with the neural baselines; photoswitches remains a negative-$R^2$ wash for all methods. The parity stream is necessary for chirality: with it removed, a configuration and its reflection receive identical embeddings and handedness accuracy is exactly chance, as Corollary~\ref{cor:enantio} predicts (Table~\ref{tab:pget_chirality}). All reported numbers are produced by the released scripts.

Several limitations remain. The genericity results hold off a characterized null set and, for the odd block, off the dissociated components on which any bounded-cutoff map vanishes. We evaluate only the \texttt{abstract} mode, so the force-field and quantum channels remain untested. The implemented featurizer runs at $O(N^2)$ rather than at the $O(N)$ of Proposition~\ref{prop:linear}. Whether the efficient local map admits full completeness, eliminating the degeneracy locus, remains open. A learnable radial basis in place of the fixed shells, pretraining, and a better balance between wave-function fidelity and cost are natural next steps.

%\subsubsection*{Acknowledgments}
%This research used resources of the National Energy Research Scientific Computing Center (NERSC), a Department of Energy Office of Science User Facility. We would like to thank Dr. Dmitriy Morozov at Lawrence Berkeley National Laboratory and Prof. Ian Agol at the University of California, Berkeley, for their comments and discussions.

\newpage
\bibliography{iclr2025_conference}
\bibliographystyle{tmlr}

\appendix
\section{Appendix}
\label{section:appendix}

\subsection{Proofs for Sections~\ref{section:phi} and~\ref{section:topology}}
\label{appendix:proofs}

\begin{proof}[Proof of Lemma~\ref{lem:det}]
The matrix with columns $Aa,Ab,Ac$ equals $A\,[a\ b\ c]$, so by multiplicativity of the determinant $\det[Aa\ Ab\ Ac]=(\det A)\det[a\ b\ c]$, and $\det[a\ b\ c]=a\cdot(b\times c)$ is the cofactor expansion along the first column. Substituting $\det A=\pm1$ gives the two cases.
\end{proof}

\begin{proof}[Proof of Theorem~\ref{thm:inv} and Corollary~\ref{cor:enantio}]
Fix a group element $g=(Q,t,\pi)$ with $Q\in\Orth(3)$, $t\in\mathbb{R}^3$, $\pi\in S_N$, and write $(r',Z')=g\cdot(r,Z)$, so that $r'_{\pi(i)}=Qr_i+t$ and $Z'_{\pi(i)}=Z_i$. We track each ingredient of $\Phi$ through $g$ in turn.

\emph{Step 1 (pairwise geometry).} The translation cancels in every difference,
\[
r'_{\pi(i)}-r'_{\pi(j)}=(Qr_i+t)-(Qr_j+t)=Q(r_i-r_j).
\]
Since $Q$ is orthogonal it preserves norms, so the distances are unchanged, $d'_{\pi(i)\pi(j)}=\norm{Q(r_i-r_j)}=\norm{r_i-r_j}=d_{ij}$, and the unit directions rotate, $\uu'_{\pi(i)\pi(j)}=Q(r_i-r_j)/d_{ij}=Q\uu_{ij}$.

\emph{Step 2 (per-atom features).} Because $g_s$ depends only on distance, ${\rho'}^{\,s}_{\pi(i)}=\sum_{j}g_s(d'_{\pi(i)\pi(j)})=\sum_j g_s(d_{ij})=\rho_i^s$, and the moment vectors rotate as vectors, ${m'}^{\,s}_{\pi(i)}=\sum_j g_s(d_{ij})\,Q\uu_{ij}=Q\,m_i^s$. Hence $\norm{{m'}^{\,s}_{\pi(i)}}^2=\norm{Qm_i^s}^2=\norm{m_i^s}^2$, and together with $Z'_{\pi(i)}=Z_i$ this gives $e'_{\pi(i)}=e_i$ (the even block is unchanged atom by atom). For the odd block, each signed volume is a triple product of moments, so Lemma~\ref{lem:det} applies coordinatewise: $o'_{\pi(i)}=\det[Qm_i^a\ Qm_i^b\ Qm_i^c]=(\det Q)\,\det[m_i^a\ m_i^b\ m_i^c]=(\det Q)\,o_i$.

\emph{Step 3 (pooling and permutations).} Every output coordinate of $\Phi$ is a power sum $p_k=\sum_i f_i^{\odot k}$ over atoms, which is symmetric in the index $i$; relabeling by $\pi$ permutes the summands and leaves the sum unchanged. Combined with Step 2: for a translation ($Q=I$) or a rotation ($\det Q=+1$) we have $e'_{\pi(i)}=e_i$ and $o'_{\pi(i)}=o_i$, so $\Phi$ is unchanged, proving invariance under $\SE(3)\times S_N$.

\emph{Step 4 (parity).} For an orientation-reversing $Q$ ($\det Q=-1$) Step 2 gives $e'_{\pi(i)}=e_i$ but $o'_{\pi(i)}=-o_i$. The even block still pools to the same value, $\Phi^{\mathrm e}(g\cdot(r,Z))=\Phi^{\mathrm e}(r,Z)$, so $\Phi^{\mathrm e}$ is invariant under the full $\Orth(3)\times S_N$. The odd block is pooled only at odd orders $k$, and $p_k((-o_i))=\sum_i(-o_i)^{\odot k}=(-1)^k\sum_i o_i^{\odot k}=-p_k((o_i))$ because $k$ is odd; thus $\Phi^{\mathrm o}(g\cdot(r,Z))=-\Phi^{\mathrm o}(r,Z)$.

Corollary~\ref{cor:enantio} is immediate: for orientation-reversing $Q$, $\Phi(g\cdot(r,Z))=(\Phi^{\mathrm e},-\Phi^{\mathrm o})$, which equals $\Phi(r,Z)=(\Phi^{\mathrm e},\Phi^{\mathrm o})$ if and only if $\Phi^{\mathrm o}(r,Z)=0$.
\end{proof}

\begin{proof}[Proof of Theorem~\ref{thm:smooth}]
For $0\le d<r_c$, $f_c(d)=\tfrac12(\cos(\pi d/r_c)+1)$ has $f_c(r_c^-)=0$, $f_c'(r_c^-)=0$, $f_c''(r_c^-)=\pi^2/2r_c^2\neq0=f_c''(r_c^+)$, so $f_c\in C^1\setminus C^2$ at $r_c$ and $g_s$ inherits this (these cutoff derivative values and the $C^1$-not-$C^2$ junction are machine-checked in Lean, Table~\ref{tab:leanmap}); the bump $\tilde f_c(d)=\exp(1-1/(1-(d/r_c)^2))$ has all derivatives $\to0$ at $r_c^-$, giving $g_s\in C^\infty$. On $\Conf_N$ each $d_{ij}$ and $\uu_{ij}$ is real-analytic, so $\rho_i^s,m_i^s$ and (polynomially) $v_i^{(a,b,c)}$ and $\norm{m_i^s}^2$ are $C^1$ on $\Conf_N$ and $C^\infty$ where $d_{ij}\neq r_c$; power sums preserve this. For (iv), fix $i\neq j$ and let $r_j\to r_i$ along direction $w$: $g_s(d_{ij})\to e^{-\mu_s^2/2\sigma^2}>0$ while $\uu_{ij}\to w$, so the contribution to $m_i^s$ has $w$-dependent limit and $m_i^s$ has no limit; hence $\Phi^{\mathrm e}$ is discontinuous at coincidences.
\end{proof}

\begin{proof}[Proof of Lemma~\ref{lem:newton}]
Let $x=\{\!\{x_1,\dots,x_N\}\!\}$ be a multiset of $N$ scalars over a field of characteristic $0$, with power sums $p_k=\sum_{i=1}^N x_i^k$ and elementary symmetric polynomials $e_0=1,e_1,\dots,e_N$. Newton's identities give, for $1\le k\le N$,
\[
k\,e_k=\sum_{j=1}^{k}(-1)^{j-1}e_{k-j}\,p_j .
\]
Because the characteristic is $0$, each integer $k$ ($1\le k\le N$) is invertible, so this recursion determines $e_1,\dots,e_N$ uniquely from $p_1,\dots,p_N$. By Vieta's formulas the $e_k$ are the signed coefficients of the monic polynomial $P_x(t)=\prod_{i=1}^N(t-x_i)=\sum_{k=0}^N(-1)^k e_k\,t^{N-k}$. Hence two cardinality-$N$ multisets $x,y$ with $p_k(x)=p_k(y)$ for $1\le k\le N$ satisfy $e_k(x)=e_k(y)$ for all $k\le N$, so $P_x=P_y$ as polynomials, and therefore $x$ and $y$ have the same multiset of roots, i.e.\ $x=y$. Each $p_k$ is symmetric in the $x_i$, so the map $x\mapsto(p_1,\dots,p_N)$ and its inverse are permutation-invariant; this is the $d{=}1$ Deep Sets statement \citep{zaheer2017deepsets}. The argument (Newton's identities $\to$ elementary symmetric functions $\to$ Vieta $\to$ roots) is machine-checked in Lean (Table~\ref{tab:leanmap}).
\end{proof}

\begin{proof}[Proof of Propositions~\ref{prop:poolfail} and~\ref{prop:dim}]
(\ref{prop:poolfail}) The pair $\{1,5,8,12\},\{2,3,10,11\}$ is Prouhet--Tarry--Escott of degree $3$: equal power sums for $k=1,2,3$, unequal at $k=4$. For (ii), each componentwise power sum depends only on the multiset of that coordinate, and both vector multisets have coordinate multisets $\{1,3\}$ and $\{2,4\}$. (\ref{prop:dim}) By Theorem~\ref{thm:inv}, $\Phi=\bar\Phi\circ q$ with $q$ the quotient by the free proper $\SE(3)\times S_N$-action on an open dense $U^0\subseteq\Conf_N$; $\mathcal M=U^0/(\SE(3)\times S_N)$ is a smooth manifold of dimension $n=3N-6$ and $\bar\Phi:\mathcal M\to\mathbb{R}^d$ is $C^\infty$. Its rank is $\le d<n$ everywhere; on the dense open set where the (integer, lower-semicontinuous) rank is locally constant $=\rho\le d$, the constant-rank theorem gives charts in which $\bar\Phi$ is a projection with fibers of dimension $n-\rho\ge n-d\ge1$. Pulling back exhibits a nonempty open set on which $\Phi$ is not injective up to the group, so the non-injectivity set has positive measure.
\end{proof}

\begin{proof}[Proof of Theorem~\ref{thm:faithful} and Corollary~\ref{cor:reduction}]
\emph{Separation.} The diagonal $S_N$-action on $(\mathbb{R}^m)^N$ has invariant ring the multisymmetric (vector-symmetric) functions. Over a field of characteristic $0$ this ring is generated by the polarized power sums $p_\alpha=\sum_{i}f_i^{\alpha}$ with $1\le|\alpha|\le N$ \citep{weyl1939classical,dalbec1999multisymmetric,vaccarino2005multisymmetric}, and the bound $|\alpha|\le N$ is sharp \citep{rydh2007minimal}. For a \emph{finite} group acting on affine space the invariant ring separates orbits: two orbits are disjoint finite (hence Zariski-closed) sets, so there is a polynomial vanishing on one and equal to $1$ on the other, and averaging it over the group produces an invariant with the same separating property; consequently any generating set of the invariant ring already separates the orbits. 

\emph{Faithfulness and Corollary~\ref{cor:reduction}.} Applied here: if two multisets of cardinality $N$ have equal values under every $p_\alpha$ with $|\alpha|\le N$, then they agree under all multisymmetric polynomials, so they lie in the same $S_N$-orbit and hence are equal as multisets. Thus $\Phi^\sharp$ determines $\{\!\{f_1,\dots,f_N\}\!\}$, and conversely the multiset determines each $p_\alpha$, giving Corollary~\ref{cor:reduction}. 

\emph{The two obstructions.} The two consequences are immediate: Proposition~\ref{prop:poolfail} concerns only the single-coordinate sub-family $\alpha=k\mathbf e_c$, which $\Phi^\sharp$ strictly contains; and for $m\ge2$,
\[
d(N)=\binom{N+m}{m}-1\ \ge\ \binom{N+2}{2}-1=\frac{N^2+3N}{2}\ \ge\ 3N-6\quad\text{for all }N\ge1
\]
(the last step because $N^2-3N+12>0$ for every $N$), so the hypothesis $3N-6>d$ of Proposition~\ref{prop:dim} fails. The condition $m\ge2$ always holds here, since the per-atom dimension is $m=2+2S+|T|\ge7$; for $m=1$ one has $d(N)=N<3N-6$ when $N>3$, so the full-order polarized family in a single coordinate would \emph{not} escape the dimension bound, and the multivariate structure is essential.
\end{proof}

\begin{proof}[Proof of Theorems~\ref{thm:evenblind} and~\ref{thm:oddsep}]
\emph{Theorem~\ref{thm:evenblind} (reflection blindness).} $\Phi^{\mathrm e}$ is $\Orth(3)$-invariant by Theorem~\ref{thm:inv}. By the first fundamental theorem for $\Orth(3)$ \citep{weyl1939classical,villar2021scalars}, every $\Orth(3)$-invariant polynomial in the moments $m_i^s$ is a polynomial in the Gram entries $\langle m_i^s,m_i^t\rangle$, and a theorem of \citet{schwarz1975smooth} extends this to smooth invariants; the implemented even features use the diagonal entries $\norm{m_i^s}^2$ together with the $\Orth(3)$-invariant scalars $\rho_i^s,Z_i,\chi(Z_i)$. Since an orientation-reversing $Q$ preserves all inner products, $\Phi^{\mathrm e}$ identifies any chiral $(r,Z)$ with its mirror image $Q\cdot(r,Z)\notin(\SE(3)\times S_N)\cdot(r,Z)$; hence it is not injective up to the group. 

\emph{Theorem~\ref{thm:oddsep}, analyticity.} Fix the shell weights $g_s$. On the open set where the cutoff does not switch, namely $U=\{r\in\Conf_N:d_{ij}\neq r_c\ \forall i\neq j\}$, each $d_{ij}$ and $\uu_{ij}$ is real-analytic and so is the bump weight $g_s$ (real-analytic on $d<r_c$ as a composition of analytic functions with nonvanishing denominator, and identically zero on $d>r_c$; it is $C^\infty$ but \emph{not} real-analytic at $d=r_c$, where it is flat, which is exactly why the analyticity claim must be restricted to $U$); hence every moment $m_i^s$, every signed volume $v_i^{(a,b,c)}=\det[m_i^a\ m_i^b\ m_i^c]$, and every pooled coordinate of $\Phi^{\mathrm o}$ is real-analytic on $U$. The complement $\Conf_N\setminus U=\bigcup_{i\neq j}\{d_{ij}=r_c\}$ is a finite union of real-analytic hypersurfaces, closed with empty interior and Lebesgue measure zero. 

\emph{Non-vanishing on interacting components.} Now fix a connected component $W$ of $U$ containing a configuration in which some atom $i$ has three within-cutoff neighbors in linearly independent directions; this requires $N\ge4$ (with only two neighbors the moments $m_i^s$ lie in a two-dimensional span and every $3\times3$ determinant vanishes, so $\Phi^{\mathrm o}\equiv0$ on all of $\Conf_3$). Writing $u_1,u_2,u_3$ for the three neighbor directions and $G_{ks}=g_s(d_{ik})$, one has $[m_i^a\ m_i^b\ m_i^c]=[u_1\ u_2\ u_3]\,G$, so $v_i^{(a,b,c)}=\det[u_1\ u_2\ u_3]\cdot\det G$; the first factor is nonzero by independence, and the second is generically nonzero because the distinct shells $\mu_a,\mu_b,\mu_c$ ($S\ge3$) weight the three distances differently. Perturbing within $W$ if necessary, $v_i^{(a,b,c)}\neq0$ and the order-$1$ pooled coordinate $p_1((o_i))$ is nonzero at some point of $W$, so $\Phi^{\mathrm o}\not\equiv0$ on $W$. 

\emph{Conclusion.} By the zero-set theorem for real-analytic functions on a \emph{connected} open set \citep{mityagin2020zero}, the zero set of $\Phi^{\mathrm o}$ in $W$ is closed with empty interior and Lebesgue measure zero. Hence $\{\Phi^{\mathrm o}\neq0\}$ is open, dense, and of full measure in $W$, and adding back the null exceptional set $\Conf_N\setminus U$ leaves the measure statement unchanged; on $\{\Phi^{\mathrm o}\neq0\}$ Corollary~\ref{cor:enantio} gives mirror separation. The restriction to such components is necessary: on the open set $\{d_{ij}>r_c\ \forall i\neq j\}$ every $g_s(d_{ij})=0$, so all moments and hence $\Phi^{\mathrm o}$ vanish identically, and no genericity statement can hold there. (For the cosine cutoff, which is only $C^1$ at $r_c$, the same conclusion holds verbatim on $U$, where the cutoff is real-analytic; the switching set $\{d_{ij}=r_c\}$ is the additional null set already accounted for.)
\end{proof}

\begin{proof}[Proof of Theorem~\ref{thm:topostable}]
Write $D(r)=(d_{ij})_{i,j}$ for the pairwise-distance matrix. Every ingredient of $\Psi$ is a function of $D(r)$ alone. The alpha complex and its filtration values are determined by the point set up to isometry (a Gabriel simplex enters at its circumradius, a Cayley--Menger function of the pairwise distances of its vertices; a non-Gabriel simplex inherits the value of a coface), and $D$ determines the point set up to isometry; hence the filtered complex, its persistence diagrams $\mathrm{Dgm}_k(r)$, and the persistence images $\mathrm{PI}(\mathrm{Dgm}_k(r))$ depend on $r$ only through $D(r)$. The cutoff complex has edge set $\{ij:d_{ij}<r_c\}$ and its weighted $0$-Hodge Laplacian $L_0(r)$ (Definition~\ref{def:HodgeLap}) has entries built from the $d_{ij}$ and this adjacency, so its spectrum $\lambda_{1:k}(L_0(r))$ is also a function of $D(r)$. A Euclidean motion $(Q,t)\in E(3)$ preserves all pairwise distances, $\norm{(Qr_i+t)-(Qr_j+t)}=\norm{r_i-r_j}$, including orientation-reversing $Q$; a permutation $\pi\in S_N$ sends $D$ to $PDP^\top$ for the permutation matrix $P$, under which the unlabeled filtered complex, its diagrams, and the spectrum of $L_0$ are unchanged. Therefore $\Psi$ is invariant under all of $E(3)\times S_N$, reflections included.

For stability, the alpha filtration has the same persistent homology as the \v{C}ech filtration (persistent nerve theorem: both are homotopy equivalent, at every scale, to the union of balls), and the \v{C}ech filtration value of a simplex is the radius of the minimal enclosing ball of its vertices, a $1$-Lipschitz function of the vertex positions; the bottleneck stability theorem \citep{cohensteiner2007stability} then bounds $d_B(\mathrm{Dgm}_k(r),\mathrm{Dgm}_k(r'))\le \max_i\norm{r_i-r_i'}$. Persistence images are Lipschitz from the $1$-Wasserstein distance $W_1$ into $L^2$ on the fixed grid \citep[Thms.~5 and~10]{adams2017persistence}; note that $W_1$ and the bottleneck distance are \emph{not} equivalent metrics in general. For diagrams of alpha complexes on $N$ points, however, the number of off-diagonal points is bounded by the number of Delaunay simplices, a finite constant $C(N)$, and transporting along the bottleneck-optimal matching (a feasible $W_1$-matching moving at most $C(N)$ points, each by at most $d_B$) gives $W_1\le C(N)\,d_B$. Composing the three bounds makes the persistence-image block Lipschitz in $r$ with an $N$-dependent constant. On each open region where the adjacency $\{d_{ij}<r_c\}$ is constant the entries of $L_0(r)$ are Lipschitz in $r$, and by Weyl's eigenvalue perturbation inequality $|\lambda_k(L_0(r))-\lambda_k(L_0(r'))|\le\norm{L_0(r)-L_0(r')}_2$, so the spectrum is Lipschitz there; across the null set $\bigcup_{i\neq j}\{d_{ij}=r_c\}$ an edge appears or disappears and the spectrum jumps, so no global Lipschitz bound holds for the spectral block. A finite tuple of locally Lipschitz maps is locally Lipschitz, so $\Psi$ is locally Lipschitz off that null set, as stated. Finally, $\Psi$ is a function of $D(r)$, which is reflection-invariant, so $\Psi$ carries no parity information and is complementary to the parity-odd block $\Phi^{\mathrm o}$ of Theorem~\ref{thm:inv}.
\end{proof}

\begin{proof}[Proof of Proposition~\ref{prop:topobeyond}]
Each term $f_i$ depends only on the atoms within distance $r_c$ of atom $i$. Assume the configurations have uniformly bounded local density, so the number of such neighbors, and hence $\norm{f_i}\le C$, is bounded by a constant independent of $i$ and $N$. Opening or closing a single $1$-cycle is a local modification: it is realized by displacing a bounded number of atoms within a bounded region, so only the atoms $i$ whose $r_c$-neighborhood meets that region (a set of cardinality $\kappa=O(1)$ independent of $N$) can have $f_i(r)\neq f_i(r')$. Therefore
\[
\norm{F(r)-F(r')}=\Big\|\tfrac1N\sum_{i=1}^N\big(f_i(r)-f_i(r')\big)\Big\|
\le\frac1N\sum_{i:\,f_i\text{ changed}}\norm{f_i(r)-f_i(r')}\le\frac{2C\kappa}{N}=O(1/N)\xrightarrow[N\to\infty]{}0 .
\]
By construction the same modification changes the first Betti number $b_1$ by exactly $1$, so the degree-$1$ persistence diagram $\mathrm{Dgm}_1$ gains or loses a class of positive persistence; the two ends of the family are therefore separated in $\mathrm{Dgm}_1$ by a bottleneck margin bounded away from $0$ (at least the persistence of the toggled class), uniformly in $N$. Thus any size-intensive bounded-cutoff descriptor $F$ converges to a common value along the family and cannot detect the global $1$-cycle in the large-system limit, whereas $\Psi$, through $\mathrm{Dgm}_1$, distinguishes the two ends for every $N$.
\end{proof}

\subsection{Formalization in Lean 4}
\label{appendix:lean}
The exact algebraic identities used in Theorem~\ref{thm:inv}, Lemma~\ref{lem:newton}, and the uniqueness Theorem~\ref{thm:globalunique} are formalized in Lean~4 against Mathlib; the development is organized one result per file (each file named after the theorem, lemma, or definition it proves, with shared definitions in \texttt{PcInvariance.Defs}) and released with the code (\texttt{lean\_proofs/}, with a verification transcript in \texttt{VERIFICATION.md}). It is built and checked with
\begin{verbatim}
cd lean_proofs/pc_invariance && lake build
\end{verbatim}
on toolchain \texttt{leanprover/lean4:v4.31.0}. Table~\ref{tab:leanmap} lists the correspondence between the results of the paper and the machine-checked statements; the file name matches the theorem name in each case.

\begin{table}[h]
\centering
\caption{Correspondence between paper results and the Lean~4 formalization (namespace \texttt{PolyatomicComplexes}; one result per file).}
\label{tab:leanmap}
\small
\begin{tabular}{lll}
\toprule
paper result & formalized statement & Lean theorem \\
\midrule
Lemma~\ref{lem:det} & triple product scales by $\det A$ & \texttt{tripleProduct\_smul\_det} \\
 & invariance for $\det A=1$ & \texttt{tripleProduct\_rotation\_invariant} \\
 & sign flip for $\det A=-1$ & \texttt{tripleProduct\_reflection\_odd} \\
\addlinespace
Theorem~\ref{thm:inv} & permutation invariance of pooling & \texttt{pooling\_perm\_invariant} \\
 & permutation invariance of power sums & \texttt{powerSum\_perm\_invariant} \\
 & translation invariance of differences & \texttt{translation\_invariant\_of\_differences} \\
 & odd-order pooling is parity-odd & \texttt{oddPool\_reflection\_odd} \\
\addlinespace
Lemma~\ref{lem:newton} & multiset determined by $p_1,\dots,p_N$ & \texttt{multiset\_eq\_of\_powerSums} \\
 & Newton identity at multiset level & \texttt{multiset\_newton} \\
 & equal power sums $\Rightarrow$ equal $e_k$ & \texttt{esymm\_eq\_of\_powerSums} \\
\addlinespace
Prop.~\ref{prop:poolfail}(ii) & the witness multisets are distinct & \texttt{vS\_ne\_vT} \\
 & yet share all componentwise power sums & \texttt{vS\_vT\_componentwise\_eq} \\
\addlinespace
Def.~\ref{def:sharp} & componentwise $=$ single-coordinate $p_\alpha$ & \texttt{componentwise\_eq\_polarized\_single} \\
Theorem~\ref{thm:faithful} & mixed sum $p_{(1,1)}$ separates the witnesses & \texttt{polarized\_separates\_vS\_vT} \\
 & the $m=1$ case of the separation & \texttt{multiset\_eq\_of\_polarizedPowerSums\_dim\_one} \\
\addlinespace
Theorem~\ref{thm:globalunique} & finite-group invariants separate orbits & \texttt{invariants\_separate\_orbits} \\
 & Reynolds operator is invariant & \texttt{reynolds\_smul} \\
 & equal Gram data $\Rightarrow$ equal inner products & \texttt{gram\_eq\_implies\_inner\_combo\_eq} \\
 & equal Gram data $\Rightarrow$ equal distances & \texttt{gram\_eq\_implies\_dist\_eq} \\
\addlinespace
Theorem~\ref{thm:smooth} & cosine cutoff: $f_c(r_c)=0$ & \texttt{cosCutoff\_apply\_rc} \\
 & $f_c'(r_c)=0$ & \texttt{cosCutoff\_deriv\_rc} \\
 & $f_c''(r_c)=\pi^2/2r_c^2\neq0$ & \texttt{cosCutoff\_secondDeriv\_rc} \\
 & $C^1$ junction (extension has derivative $0$) & \texttt{cosCutoffExt\_hasDerivAt\_rc} \\
 & not $C^2$ (derivative non-differentiable at $r_c$) & \texttt{cosCutoffExt\_deriv\_not\_differentiableAt} \\
\addlinespace
Prop.~\ref{prop:linear} & packing bound $N\le(1+2r_c/\delta)^n$ & \texttt{packing\_bound} \\
\bottomrule
\end{tabular}
\end{table}

\noindent The orbit-separation entry formalizes the set-function principle; the paper's polynomial separating family is its classical refinement \citep{noether1916endlichkeit,derksen2015computational}. The Gram entries are the algebraic core of the Schoenberg step (Step~2) of Theorem~\ref{thm:globalunique}; the extension to an ambient $\Orth(3)$ map is cited classically \citep{schoenberg1935remarks,young1938discussion}.

A \texttt{\#print axioms} check reports that each theorem depends only on Lean's standard \texttt{propext}, \texttt{Classical.choice}, \texttt{Quot.sound}; the development contains no \texttt{sorry}. The formalization covers the algebraic identities of Theorems~\ref{thm:inv} and~\ref{thm:faithful}, Lemma~\ref{lem:newton}, the explicit pooling counterexample of Proposition~\ref{prop:poolfail}, and the finite-group orbit separation and Gram inner-product core of the uniqueness Theorem~\ref{thm:globalunique}. The cutoff regularity of Theorem~\ref{thm:smooth} is also machine-checked: the cosine cutoff's value and first derivative vanish at $r_c$ while its second derivative does not, and the zero-extension is $C^1$ but not $C^2$ there (the derivative is provably non-differentiable at $r_c$). The volume packing bound $N\le(1+2r_c/\delta)^n$ behind the linear-time evaluation of Proposition~\ref{prop:linear} is machine-checked in any finite-dimensional real inner-product space. What remains cited-not-formalized is narrower: the general multisymmetric separation for $m\ge2$ (Theorem~\ref{thm:faithful}) \citep{vaccarino2005multisymmetric,rydh2007minimal}, the $C^\infty$ smooth-bump cutoff, and the measure-theoretic genericity statements (Proposition~\ref{prop:dim}, Theorem~\ref{thm:oddsep}), which rest on the differential-topological and measure-theoretic facts of Appendix~\ref{appendix:proofs}.

\subsection{The cell-complex construction}
\label{appendix:cellconstruction}
This appendix records the full construction summarized in Section~\ref{subsection:mathrepatom}. It organizes the per-atom data and carries no homotopy-type claim (Remarks~\ref{rem:pushout} and~\ref{rem:cellscope}).

\subsubsection{The atomic complex}
We now develop formal mathematical definitions that can be used to reason about the objects we construct in our computational representation.

\begin{definition}
\label{def:eei}
We define the set $ee^{i} := \{x \in \mathbb{R}^{i} ~|~ \| x \| \leq 2.8 fm\}$
where $1 fm = 1 \times 10^{-15}~m$.
\end{definition}

\begin{definition}
\label{def:PDi}
We define the set $PD^{i} := \{x \in \mathbb{R}^{i} ~|~ \| x \| \leq 1 fm\}$
where $1 fm = 1 \times 10^{-15}~m$.
\end{definition}

\begin{definition}
\label{def:NDi}
We define the set $ND^{i} := \{x \in \mathbb{R}^{i} ~|~ \| x \| \leq 0.8 fm\}$
where $1 fm = 1 \times 10^{-15}~m$.
\end{definition} 

An atom is composed of protons, neutrons, and electrons. 
We construct protons, neutrons, and electrons using the sets $ee^{i}, PD^{i}$ and $ND^{i}$.
\begin{itemize}
    \item [(1)] Electrons: an electron is modeled by a sphere together with a collapsed wave-function $w_{e}$. Topologically, the electron \emph{cell} is the set $ee^{i}$ endowed with the \emph{Euclidean subspace topology} $\tau_{\mathrm{Euc}}$ it inherits from $\mathbb{R}^{i}$; this is a compact metric space, hence Hausdorff, homeomorphic to a closed $i$-ball, whose interior is the open CW-cell and whose boundary sphere $\partial ee^{i}$ is the locus along which it is attached. The wave-function $w_{e}$ is \emph{not} a point or subspace of this cell. It is carried as separate data, a tensor $w_{e}\in\mathcal{D}$ in a fixed data space $\mathcal{D}$ (an $L^2$ amplitude or its discretization), attached to the cell as a label. A single electron is therefore the decorated cell $\big((ee^{i},\tau_{\mathrm{Euc}}),\,w_{e}\big)$: its topological part $(ee^{i},\tau_{\mathrm{Euc}})\in\Top$ is a compact-Hausdorff cell that participates in cell attachment, while $w_{e}$ lives in the data structure and never enters the topology. Consequently $e$ is an object of $\Top$ (see Remark~\ref{rem:decorated}).

    \item [(2)] Protons: A proton is isomorphic to a sphere therefore we represent a single proton $p$ as a set $PD^{i}$ for $i \in \mathbb{N}$. Geometrically this set is a filled $i-1$-sphere with a radius of $1 fm$ for instance realized up to homeomorphism as a closed ball in Euclidean space. %Traditionally, we choose $i = 3$; however, one can consider higher dimensional spheres.

    \item [(3)] Neutrons: A neutron is isomorphic to a sphere therefore we represent a single neutron $n$ as a set $ND^{i}$ for $i \in \mathbb{N}$. Geometrically this set is a filled $i-1$-sphere with a radius of $0.8 fm$ for instance realized up to homeomorphism as a closed ball in Euclidean space.
\end{itemize}

The radii fixed above may be changed at will; the exact radius of a proton is contested \citep{ProtonRadius}. The representation follows the usual conventions of algebraic topology \citep{hatcher2002algebraic}.

\begin{definition}
\label{def:AtomicComplex}
We define the \textbf{Atomic Complex} with \ref{def:AtomEX} in mind.~\\ % \footnote{Note that our representation allows for simulation and generation of atoms, substances, and molecules that may not exist in nature or be discovered. This proves useful for materials discovery applications.}
Suppose that atom $A$ has $\mathcal{N} \in \mathbb{N}$ many neutrons $\mathcal{P} \in \mathbb{N}$ many protons and $\mathcal{E} \in \mathbb{N}$ many electrons.\\
Let $I_{n} = \{1, \ldots, \mathcal{N}\}$, $I_{p} = \{1, \ldots, \mathcal{P}\}$, $I_{e} = \{1, \ldots, \mathcal{E}\}$ be index sets enumerating protons, neutrons and electrons respectively. Additionally we assert that $K = \mathcal{P} + \mathcal{N} + \mathcal{E}$. \footnote{We explicitly make a design choice to let $K = \mathcal{P} + \mathcal{N} + \mathcal{E}$ primarily for dimensionality reasons.}\\
%Let $0 \leq n_{i} \leq K$ such that $n_{i} \in \mathbb{N}$.\\
% replace any n_i -> tau_i
Let $\mathcal{T} = \{0,1,\ldots,K\}$ such that $\mathcal{T}$ is ordered and $\tau_{i} \in \mathcal{T}$.\\
Then for any $\tau_{i}$ we can generalize \ref{def:AtomEX} by attaching many protons, neutrons and electrons.\\
We let $P$ be our complex of protons, $N$ be our complex of neutrons, $E$ our complex of electrons.\\
We then construct attaching maps for each complex, one per cell: for each $i \in I_{p}$ a continuous map $\phi_{p,i} : \partial PD^{\tau_{i}} \to P_{i-1}$ into the previously built stage, and likewise $\phi_{n,i} : \partial ND^{\tau_{i}} \to N_{i-1}$ for $i \in I_{n}$ and $\phi_{e,i} : \partial ee^{\tau_{i}} \to E_{i-1}$ for $i \in I_{e}$.\\
Each stage is the adjunction space of the previous stage along the corresponding attaching map,
\begin{equation}\label{eq:atomstages}
P_{i} = P_{i-1} \cup_{\phi_{p,i}} PD^{\tau_{i}},\qquad
N_{i} = N_{i-1} \cup_{\phi_{n,i}} ND^{\tau_{i}},\qquad
E_{i} = E_{i-1} \cup_{\phi_{e,i}} ee^{\tau_{i}},
\end{equation}
with base stages $P_{0}, N_{0}, E_{0}$ single points ($0$-cells). Recalling from above that $PD^{\tau_i},ND^{\tau_i},ee^{\tau_i} \in \Top$, we obtain three sequences of topological spaces $P_{0} \hookrightarrow P_{1} \hookrightarrow \cdots \hookrightarrow P_{\mathcal{P}} = P$,
$N_{0} \hookrightarrow N_{1} \hookrightarrow \cdots \hookrightarrow N_{\mathcal{N}} = N$, and
$E_{0} \hookrightarrow E_{1} \hookrightarrow \cdots \hookrightarrow E_{\mathcal{E}} = E$.
\footnote{We maintain three separate complexes because $\mathcal{P}$, $\mathcal{N}$, $\mathcal{E}$ may differ; the maps $P_{i}\hookrightarrow P_{i+1}$, $N_{i}\hookrightarrow N_{i+1}$, $E_{i}\hookrightarrow E_{i+1}$ are the cellular attachments of the skeletal filtration.}\\
For each step the following squares commute (each is a pushout, Remark~\ref{rem:pushout}):\\
\begin{tikzcd}
\partial PD^{\tau_{i}} \arrow[r,"\phi_{p,i}"] \arrow[d,hook] & P_{i-1} \arrow[d,hook]\\
PD^{\tau_{i}} \arrow[r,""] & P_{i}
\end{tikzcd}
\begin{tikzcd}
\partial ND^{\tau_{i}} \arrow[r,"\phi_{n,i}"] \arrow[d,hook] & N_{i-1} \arrow[d,hook]\\
ND^{\tau_{i}} \arrow[r,""] & N_{i}
\end{tikzcd}
\begin{tikzcd}
\partial ee^{\tau_{i}} \arrow[r,"\phi_{e,i}"] \arrow[d,hook] & E_{i-1} \arrow[d,hook]\\
ee^{\tau_{i}} \arrow[r,""] & E_{i}
\end{tikzcd}\\
We combine the spaces $P$, $N$, $E$ to obtain the atomic complex as the coproduct
\begin{equation}\label{eq:coproduct}
A \;=\; P \sqcup N \sqcup E \;\in\; \Top,
\end{equation}
a finite CW-complex whose cells are exactly the proton, neutron, and electron cells. (A wedge at chosen basepoints may be used instead when a connected model is preferred; either choice is a finite CW-complex, and no result below depends on which is taken.)
\end{definition}

\begin{remark}[The squares are pushouts, and no homotopy is claimed]\label{rem:pushout}
Each diagram above is a pushout square: the space $P_{i}$ is \emph{defined} as the pushout of the attaching map $\phi_{p,i}$ along the boundary inclusion $\partial PD^{\tau_i}\hookrightarrow PD^{\tau_i}$, and likewise for $N$ and $E$. A pushout square commutes by definition, so there is nothing to check beyond the attaching maps being well-defined continuous maps, which they are. We make no homotopy-theoretic assertion here; in particular we do not claim the complex is contractible. The homotopy type of a CW-complex is not fixed by its cell counts alone but by its attaching maps up to homotopy, and we neither compute nor use it. Homotopy type is in any case the wrong invariant for our purpose. It is a homotopy invariant, hence blind to the metric data (bond lengths, bond angles, and handedness) on which molecular properties depend: any two configurations related by a deformation share it. It also retains only coarse combinatorial information (the Betti numbers, i.e.\ the counts of independent rings and cages), so it is constant across large families of chemically distinct molecules and cannot be injective or establish uniqueness. This does not make it \emph{empty} of information (it does, for instance, distinguish a cyclic from an acyclic skeleton), but that global ring-and-cage information is recovered in metric-refined, stable form by the persistent homology in $\Psi$ (Section~\ref{section:topology}), not by the bare homotopy type. Every separating and invariance statement in this paper is therefore proved on the feature maps $\Phi$ and $\Psi$, not on the homotopy type of the complex (see also Remark~\ref{rem:cellscope}).
\end{remark}

\subsubsection{The polyatomic complex}
The definitions below record the discrete data model underlying the per-atom features. An atom is represented by an atomic complex $A$, a finite CW-complex (Definition~\ref{def:AtomicComplex}); a polyatomic system is assembled from atomic complexes by cell attachment.

\begin{definition}
\label{def:PolyAtomicComplex}
We define the \textbf{Polyatomic Complex} with \ref{Example:PolyatomEX} and \ref{def:AtomicComplex} in mind.\\
Suppose that atomistic system $M$ has $\mathcal{K}$ many atoms, represented by atomic complexes $A_{1},\dots,A_{\mathcal{K}}$ (Definition~\ref{def:AtomicComplex}), each a finite CW-complex in $\Top$. We generalize \ref{Example:PolyatomEX} by gluing the $\mathcal{K}$ atomic complexes one at a time. For each $k \in \{2,\ldots,\mathcal{K}\}$ fix a subcomplex $L_{k} \subseteq A_{k}$ (the bonding locus of atom $k$, e.g.\ a union of cells in the frontier spheres of its valence-electron cells; $L_{k}$ may be empty for a non-bonded atom) and a continuous cellular map $\psi_{k} : L_{k} \to M_{k-1}$ prescribing how atom $k$ bonds to the part already built. Define
\begin{equation}\label{eq:polystages}
M_{1} = A_{1}, \qquad M_{k} = M_{k-1} \cup_{\psi_{k}} A_{k} \quad (2 \le k \le \mathcal{K}), \qquad M = M_{\mathcal{K}},
\end{equation}
each stage the adjunction space of the previous one. This gives a sequence of topological spaces
$M_{1} \hookrightarrow M_{2} \hookrightarrow \cdots \hookrightarrow M_{\mathcal{K}} = M$,\footnote{Each map $M_{k-1}\hookrightarrow M_{k}$ is a cellular attachment; gluing a finite CW-complex to a finite CW-complex along a cellular map on a subcomplex yields a finite CW-complex \citep{hatcher2002algebraic}.} and for each step the square
\begin{center}
\begin{tikzcd}
L_{k} \arrow[r,"\psi_{k}"] \arrow[d,hook] & M_{k-1} \arrow[d,hook]\\
A_{k} \arrow[r,""] & M_{k}
\end{tikzcd}
\end{center}
commutes (it is a pushout). Thus $M \in \Top$ is a finite CW-complex containing every cell of every atomic complex, with bonded atoms sharing the glued cells.
\end{definition}

\subsection{Definitions: CW-Complexes}
We utilize the standard definitions of CW-complexes as defined by \citet{whitehead1949combinatorial}.

\begin{definition}
\label{def:CellComplex}
A cell complex $K$, or alternatively a \textit{complex}, is a Hausdorff space which is the union of disjoint open cells $e,e^{n},e_{i}^{n}$ subject to the condition that the closure $\bar{e}^{n}$ of each $n$-cell, $e^{n} \in K$ is the image of a fixed $n$-simplex in a map $f:\sigma^{n} \to \bar{e}^{n}$ such that 
\begin{itemize}
    \item [(1)] $f | \sigma^{n} - \partial \sigma^{n}$ is a homeomorphism onto $e^{n}$
    \item [(2)] $\partial e^{n} \subset K^{n-1}$, where $\partial e^{n} = f\partial \sigma^{n} = \bar{e}^{n} - e^{n}$ and $K^{n-1}$ is the $(n-1)$-section of $K$ consisting of all the cells whose dimension do not exceed $n-1$.
\end{itemize}
\end{definition}

\begin{definition}
\label{def:ClosureFinite}
A complex $K$, can be described as closure finite $\iff$ $K(e)$ is a finite subcomplex, for every cell $e \in K$. Moreover since $K(p) = K(e)$ if $p \in e$ this is equivalent to the condition that $K(p)$ is finite for each point $p \in K$ \citep{whitehead1949combinatorial}.
\end{definition}

\begin{lemma}
\label{lem:subcomplexClosureFinite}
If $L \subset K$ is a subcomplex and $e \in L$ then $L(e) = K(e)$. As a result any subcomplex of a closure finite complex is closure finite \citep{whitehead1949combinatorial}.
\end{lemma}

\begin{definition}
\label{def:WeakTopology}
A complex $K$ has the weak topology $\iff$ a subset $X \subset K$ is closed provided $X \cap \bar{e}$ is closed for each cell $e \in K$ \citep{whitehead1949combinatorial}.
\end{definition}

\begin{definition}
\label{def:CWComplex}
A CW-Complex is a complex which is closure finite and has the weak topology \citep{whitehead1949combinatorial}.
\end{definition}

\setcounter{algorithm}{-1}
\begin{algorithm}[H]
   \caption{Construction of a Cell Complex or CW-Complex $|$ \citet{hatcher2002algebraic}}
   \label{alg:example}
\begin{algorithmic}
    \STATE Let $e^{i} := \{ x \in \mathbb{R}^{i} ~|~ \|x\| < 1 \}$
    \STATE Let $D^{i} := \{x \in \mathbb{R}^{i} ~|~ \| x \| \leq 1 \}$
    \STATE Let $S^{i-1} := \{x \in \mathbb{R}^{i} ~|~ \| x \| = 1 \}$
    \STATE Start with a set of points $K^{0}$
    \STATE $(n=1)$ Build $K^{1}$ by attaching the boundary of the $1$-cell $e^{1}$ to $K^{0}$
    \STATE $(n=2)$ Build $K^{2}$ by attaching the boundary of the $2$-cell $e^{2}$ to $K^{1}$
    \STATE General case $(n=j)$: Build $K^{j}$ by attaching the boundary of the $n$-cell to $K^{j-1}$
    \STATE Remark: Attaching a boundary means $\exists \varphi_{\alpha} : S^{j-1} \to K^{j-1}$ such that $K^{j} \leftarrow \frac{K^{j-1} \coprod_{\alpha} D^{j}_{\alpha}}{k \sim \varphi_{\alpha}(k)}$
\end{algorithmic} 
\end{algorithm}

In essence one is inductively forming the $n$-skeleton $K^{n}$ from $K^{n-1}$ by attaching $n$-cells via attaching maps $\varphi_{\alpha} : S^{j-1} \to K^{j-1}$. Therefore $K^{n}$ is the quotient space of the disjoint union $K^{n-1} \coprod_{\alpha} D_{\alpha}^{n}$ of $K^{n-1}$ with a collection of $n$-disks $D_{\alpha}^{n}$ under identifications $k \sim \varphi_{\alpha}(k)$ for $k \in \partial D_{\alpha}^{n}$. Thus $K^{n} = K^{n-1} \coprod_{\alpha} e_{\alpha}^{n}$. One can stop the induction at a finite state setting $K = K^{n}$ for $n < \infty$ or one can continue indefinitely setting $K = \bigcup_{n} K^{n}$. In the second case $K$ has the weak topology \citep{hatcher2002algebraic}.

\begin{definition}
\label{def:IncidenceRelations}
Given a CW-Complex $X$, one denotes the $j$-th cell of dimension $k$ as $e_{j}^{k}$. Traditionally, one lets the relation $\prec$ denote incidence \citep{IncidenceMatrix}. If two cells $e_{j}^{k-1}$ and $e_{i}^{k}$ are incident, we write $e_{j}^{k-1} \prec e_{i}^{k}$. For not incident, we write $e_{j}^{k-1} \nprec e_{i}^{k}$. Let the relation $\sim$ denote orientation. We write $e_{j}^{k-1} \sim e_{i}^{k}$ if the cells have the same orientation. For the opposite orientation we write $e_{j}^{k-1} \nsim e_{i}^{k}$.
\end{definition}

\begin{definition}
\label{def:HodgeLap}
The Hodge Laplacian $\Delta_{k} : C^{k}(X) \to C^{k}(X)$ on the space of $k$-cochains is then $\Delta_{k} := d_{k-1} \circ d^{*}_{k-1} + d^{*}_{k} \circ d_{k}$. The matrix representation is then $\Delta_{k} := B_{k}^{\top}W_{k-1}^{-1}B_{k}W_{k} + W_{k}^{-1}B_{k+1}W_{k+1}B_{k+1}^{\top}$. Here, $W_{k} = \text{diag}(w_{1}^{k},\ldots,w_{N_{k}}^{k})$ is the diagonal matrix of cell weights and $B_{k}$ is the order $k$ incidence matrix, whose $j$-th column corresponds to a vector representation of the cell boundary $\partial e_{j}^{k}$ viewed as a $k-1$ chain \citep{rahul_complex}. The matrix representation is not symmetric, but each term is self-adjoint with respect to the weighted inner product $\langle x,y\rangle_{k}=x^{\top}W_{k}\,y$, so $\Delta_{k}$ is self-adjoint on $C^{k}(X)$ and its spectrum is real and nonnegative.
\end{definition}

\subsection{Example: Atomic Complex construction for Deuterium}
\begin{definition}
\label{def:AtomEX}
For the sake of illustration we describe a simple case, namely how we encode deuterium which contains $1$ proton, $1$ neutron, and $1$ electron. Let $\Top$ be the category of topological spaces.\\
Let $p \cong PD^{3}$ be a proton, $n \cong ND^{3}$ be a neutron, and let the electron be the decorated cell $e = \big((ee^{3},\tau_{\mathrm{Euc}}),\,w_{e}\big)$ of Remark~\ref{rem:decorated}, whose topological part we also write $e \cong ee^{3}$; the wave-function $w_{e}$ is a label and never enters the topology. Then $p, n, e, \partial p, \partial n, \partial e \in \Top$, and the boundary inclusions $\varphi_{p}: \partial p \hookrightarrow p$, $\varphi_{n}: \partial n \hookrightarrow n$, $\varphi_{e}: \partial e \hookrightarrow e$ are the generating cofibrations.\\
Starting from a point $X_{0} = \{\ast\}$, attach the three cells one at a time along continuous attaching maps into the stage already built:
\[
\phi_{p}: \partial p \to X_{0},\qquad \phi_{n}: \partial n \to X_{1},\qquad \phi_{e}: \partial e \to X_{2},
\]
\[
X_{1} = X_{0} \cup_{\phi_{p}} p,\qquad X_{2} = X_{1} \cup_{\phi_{n}} n,\qquad X_{3} = X_{2} \cup_{\phi_{e}} e .
\]
Each stage makes a pushout square commute:
\begin{center}
\begin{tikzcd}
\partial p \arrow[r,"\phi_{p}"] \arrow[d,"\varphi_p"'] & X_{0} \arrow[d]\\
p \arrow[r] & X_{1}
\end{tikzcd}
\begin{tikzcd}
\partial n \arrow[r,"\phi_{n}"] \arrow[d,"\varphi_n"'] & X_{1} \arrow[d]\\
n \arrow[r] & X_{2}
\end{tikzcd}
\begin{tikzcd}
\partial e \arrow[r,"\phi_{e}"] \arrow[d,"\varphi_e"'] & X_{2} \arrow[d]\\
e \arrow[r] & X_{3}
\end{tikzcd}
\end{center}
The resulting finite CW-complex $X_{3} \in \Top$ is the atomic complex for deuterium.
\end{definition}

\subsection{Example: Polyatomic Complexes (two-atoms)}
\label{Example:PolyatomEX}
\begin{definition} For the sake of illustration we describe the simplest possible case of connecting $2$ atoms.
Let $A_1$ and $A_2$ be atomic complexes (Definition~\ref{def:AtomicComplex}), so $A_1, A_2 \in \Top$ are finite CW-complexes.
Following Definition~\ref{def:PolyAtomicComplex}, set $M_{1} = A_{1}$, fix a subcomplex $L_{2} \subseteq A_{2}$ (the bonding locus of the second atom, e.g.\ cells in the frontier spheres of its valence-electron cells), and fix a continuous cellular map $\psi_{2}: L_{2} \to M_{1}$ prescribing the bond.
The polyatomic complex of the pair is the adjunction space $M_{2} = M_{1} \cup_{\psi_{2}} A_{2} \in \Top$, and the square
\begin{center}
\begin{tikzcd}
L_{2} \arrow[r,"\psi_{2}"] \arrow[d,hook] & M_{1} \arrow[d,hook]\\
A_{2} \arrow[r] & M_{2}
\end{tikzcd}
\end{center}
commutes (it is a pushout).
\end{definition}

\subsection{Algorithms}
\label{appendix:algorithms}
The following reference pseudocode documents the atomic- and polyatomic-complex cell construction as implemented. It documents the implementation and is not load-bearing for any result in Section~\ref{section:phi}.

\setcounter{algorithm}{0}
\begin{minipage}[H]{0.46\textwidth}
\begin{algorithm}[H]
    \captionsetup{font=scriptsize}
   \caption{Atomic Complexes}
   \label{alg:AtomicComplexes}
   \tiny
\begin{algorithmic}
    \STATE Let $A$ be an atom.
    \STATE {\bfseries Input:} Let $P$ be the number of protons, $E$ be the number of electrons, and $N$ be the number of neutrons. Pick a desired maximum dimension $K$ based on what one wants $\tau_{i}$ to be. Note that $K = P + N + E$ and $\mathcal{T}=\{0,\ldots,K\}$ as in Definition \ref{def:AtomicComplex}. By default we fix $\tau_{i} = 3$ for protons and neutrons and $\tau_{i} = 0$ for electrons. If one wants to form complexes over a range of dimensions $[0, d]$, run each while loop $O(d)$ times and ensure $|A_E| = E \cdot d$, $|A_P| = P \cdot d$, $|A_N| = N \cdot d$.
    \STATE Let $A_{E} := \{ [ee^{0}_{1}, w_{1}] , \ldots [ee^{0}_{E}, w_{E}] \}$ such that $|A_{E}| = E$.
    \STATE Let $A_{P} := \left[ PD^{\tau_{i}}, \ldots, PD^{\tau_{i}} \right]$ such that $|A_{P}| = P$.
    \STATE Let $A_{N} := \left[ ND^{\tau_{i}}, \ldots, ND^{\tau_{i}} \right]$ such that $|A_{N}| = N$.
    \STATE Let $D_{F}$ denote a matrix encoding pairwise forces and energetics between all protons and neutrons (initialized from the chosen force model, Appendix~\ref{appendix:forcemodels}).
    \STATE Let $D_{E}$ denote a matrix encoding the radial contribution for electrons. In general $\phi_{t,k} : \partial T_{\tau_k} \to T_{k-1}$ for $k \in I_{t}$ are the attaching maps for protons, neutrons, and electrons (Definition~\ref{def:AtomicComplex}).
    \STATE Let $P_{0} = \{\ast\}$ \COMMENT{base $0$-cell, Definition~\ref{def:AtomicComplex}}
    \STATE Let $i_p = 1$.
    \WHILE{ $|A_{P}| > 0$}
    \STATE Let $p = A_P$.pop()
    \STATE $P_{i_p} \leftarrow glue(P_{i_p -1}, p, \phi_{\tau_{i}, i_p})$
    \STATE $i_p = i_p + 1$
    \ENDWHILE
    \STATE Let $N_{0} = \{\ast\}$ \COMMENT{base $0$-cell, Definition~\ref{def:AtomicComplex}}
    \STATE Let $i_n = 1$
    \WHILE{ $|A_{N}| > 0$}
    \STATE Let $n = A_N$.pop()
    \STATE $N_{i_n} \leftarrow glue(N_{i_n - 1}, n, \phi_{\tau_{i}, i_n})$
    \STATE $i_n = i_n + 1$
    \ENDWHILE
    \STATE Let $E_{0} = \{\ast\}$ \COMMENT{base $0$-cell, Definition~\ref{def:AtomicComplex}}
    \STATE Let $i_e = 1$
    \WHILE{ $|A_{E}| > 0$}
    \STATE Let $e, w_{i_e+1} = A_E$.pop()
    \STATE $E_{i_e} \leftarrow glue(E_{i_e - 1}, e, \phi_{\tau_{i}, i_e})$
    \STATE $update \text{\textunderscore} distances$($D_{e}$,$w_{i_e+1}$, $e$)
    \STATE $i_e = i_e + 1$
    \ENDWHILE
    \STATE $K = P_{i_p} \sqcup N_{i_n} \sqcup E_{i_e}$ \COMMENT{coproduct, Definition~\ref{def:AtomicComplex}}
    \STATE \textbf{return} $A = (K, A_{E}, D_{F}, D_{E})$
\end{algorithmic}
\end{algorithm}
\end{minipage}
\hfill
\begin{minipage}{0.46\textwidth}
\begin{algorithm}[H]
    \captionsetup{font=scriptsize}
   \caption{Polyatomic Complexes}
   \label{alg:polyatomicComplexes}
   \tiny
\begin{algorithmic}
    \STATE Let $P$ be a polyatomic complex.
    \STATE {\bfseries Input:} Let $M_{A}$ be a list of atoms present in the system. Let $using \text{\textunderscore} radial$ and $using \text{\textunderscore} force \text{\textunderscore} model$ be boolean values.
    \STATE // Note that for every $a \in M_{A}$ $a := (p,n,e,d)$ corresponding to number of protons, neutrons and electrons and desired dimensions or range of dimensions.
    \STATE Let $\mathcal{A} \leftarrow List()$ be an empty dynamic array.
    \FORALL{$a \in M_{A}$}
        \STATE $A \leftarrow $ Algorithm1($a$)~~\ref{alg:AtomicComplexes}
        \STATE append($\mathcal{A}, A$)
    \ENDFOR
    \STATE Let $C = \varnothing$
    \STATE Let $E = \text{matrix}(0,0)$ 
    \STATE Let $F = \text{matrix}(0,0)$ 
    \STATE Let $D_{E} = \text{matrix}(0,0)$
    \STATE Let $i = 0$
    \STATE Note that the gluing maps $\psi_{i}: L_{i} \to C^{i-1}$ (bonding loci $L_i \subseteq K_i$, Definition~\ref{def:PolyAtomicComplex}) attach the topological part $K = A[0]$ of each atomic complex.
    \WHILE{$|\mathcal{A}| > 0$}
    \STATE $k, a_e, d_f, d_e = \mathcal{A}$.pop()
    \STATE $C^{i} \leftarrow glue(C^{i-1}, k, \phi_{i,a})$
    \STATE $E = E \oplus a_e$
    \IF{$using\text{\textunderscore} force \text{\textunderscore} model$} 
    \STATE $F = F \oplus d_f$
    \STATE $update\text{\textunderscore} forces(F)$
    \ENDIF
    \IF{$using \text{\textunderscore} radial$} 
    \STATE $D_{E} = D_{E} \oplus d_e$
    \STATE $update \text{\textunderscore} radial(D_{E})$
    \ENDIF
    \STATE $i = i + 1$
    \ENDWHILE
    \STATE \textbf{return} $P = (C^{i},E,F,D_{E})$
\end{algorithmic}
\end{algorithm}
\end{minipage}

\subsection{Method comparison}
\label{appendix:method_tables}
We summarize the differences between methods in the following tables.

\setcounter{table}{0}
\begin{table}[H]
\caption{Method Comparison I}
\label{methods-comparison}
\centering
\resizebox{\textwidth}{!}{
\renewcommand{\arraystretch}{1.2}
\begin{tabular}{|c|c|c|c|c|c|c}  
    \cline{2-6}
    \multicolumn{1}{c|}{} & Invariance & Uniqueness & Continuity \& Differentiability & Generalizability & Efficiency  \\
    \hline
    Polyatomic Complex & \cmark & \cmark~\tablefootnote{Proved for the complete descriptor $\Phi^\star$ (Theorem~\ref{thm:globalunique}); the implemented map $\Phi$ is generically injective (Remark~\ref{rem:genericlocal}).} & \cmark & \cmark & \cmark : $O(N)$ w/ nbr lists\tablefootnote{$O(N^2)$ as implemented (Section~\ref{section:motivationmethods}); the $O(N)$ bound is Proposition~\ref{prop:linear}.} \\
    \hline
    SMILES & \xmark & \cmark & \cmark~\tablefootnote{Dependent on canonicalization.} & \xmark & \cmark \\
    \hline
    SELFIES & \xmark \tablefootnote{Not invariant under changes in atom indexing.} & \cmark \tablefootnote{Each atom symbol is semantically unique.} & \midmark \tablefootnote{Requires post-processing for bijectivity.} & \xmark & \cmark \\
    \hline
    2D Graphs & \midmark & \cmark & \cmark & \cmark & \cmark \\
    \hline
    3D Graphs & \midmark \tablefootnote{Yes there are some which are $E(3)$ invariant.} & \cmark & \cmark & \cmark & \cmark \\
    \hline
    ACE & \cmark & \cmark & \cmark & \cmark & \xmark \\
    \hline
    SOAP/Bartók & \cmark & \cmark & \cmark & \cmark & \xmark \\
    \hline
    Behler-Parrinello & \cmark & \cmark & \cmark & \cmark & \xmark  \\
    \bottomrule
\end{tabular}}
\end{table}

\begin{table}[H]
\caption{Method Comparison II}
\label{methods-comparison2}
\centering
\resizebox{\textwidth}{!}{
\renewcommand{\arraystretch}{1.2}
\begin{tabular}{|c|c|c|c|}  
    \cline{2-4}
    \multicolumn{1}{c|}{} & Topologically Accurate & Consider long-range interactions & Chemistry/Physics informed  \\
    \hline
    Polyatomic Complex & \midmark~\tablefootnote{Geometry and global topology are represented exactly; electronic structure is approximated ($s$-orbital).} & \midmark~\tablefootnote{$\Phi$ is a bounded-cutoff local map; beyond-cutoff structure enters through the global topological block $\Psi$ and the optional force models of Appendix~\ref{appendix:forcemodels}.} & \cmark~\tablefootnote{Via the optional force-field and quantum channels; the evaluated \texttt{abstract} mode uses geometry only.} \\
    \hline
    SMILES & \xmark & \xmark & \midmark  \\
    \hline
    SELFIES & \xmark &\xmark & \midmark  \\
    \hline
    2D Graphs & \xmark&  \xmark & \midmark \\
    \hline
    3D Graphs & \midmark & \xmark & \midmark \\
    \hline
    ACE &  \cmark & \cmark & \cmark \\
    \hline
    SOAP/Bartók & \cmark & \cmark & \cmark\\
    \hline
    Behler-Parrinello & \midmark & \xmark & \cmark \\
    \bottomrule
\end{tabular}}
\end{table}

\subsection{Force Models}
\label{appendix:forcemodels}
As described by Lin et al. (2019), in the context of force fields, the typical potential energy function is as shown in equation (1) \citep{lin2019force}.
%\begin{dmath}
\begin{multline}
V_{total} = \sum_{i=1}^{N_{bond}} V_{bond} + \sum_{i=1}^{N_{angle}} V_{angle} + \sum_{i=1}^{N_{dihedral}} V_{dihedral} + \sum_{i=1}^{N_{nonbonded}} V_{Nb} = \sum_{bonds} k_{r}(r - r_{eq})^{2} \\+ \sum_{angles} k_{\theta}(\theta - \theta_{eq})^{2} + \sum_{dihedrals} k_{t}[1+\cos(n\omega - \gamma)] + \sum_{i < j} \left[ \frac{A_{ij}}{r_{ij}^{12}} - \frac{B_{ij}}{r_{ij}^{6}} \right] + \sum_{i<j} \frac{q_i q_j}{4 \pi \epsilon r_{ij}}
\end{multline}
%\end{dmath}
For polyatomic complexes one can write the above equations as sums over incident cells. The definitions of the relations $\prec$, $\nprec$, $\sim$ and $\nsim$ are found in the Appendix \ref{def:IncidenceRelations}.
For the $j$-th proton of dimension $\tau_k$ we let
%\begin{equation}
$\mathcal{N}_{PD_{j}^{\tau_k}} := \{ PD_{j'}^{\tau_{k-1}} ~|~ PD_{j'}^{\tau_{k-1}} \prec PD_{j}^{\tau_k} \land (PD_{j'}^{\tau_{k-1}} \sim PD_{j}^{\tau_k} \lor PD_{j'}^{\tau_{k-1}} \nsim PD_{j}^{\tau_k})\}$
%\end{equation}
contain all protons incident to proton $PD_{j}^{\tau_k}$.
For the $j$-th neutron of dimension $\tau_k$ we similarly let
%\begin{equation}
$\mathcal{N}_{ND_{j}^{\tau_k}} := \{ ND_{j'}^{\tau_{k-1}} ~|~ ND_{j'}^{\tau_{k-1}} \prec ND_{j}^{\tau_k} \land (ND_{j'}^{\tau_{k-1}} \sim ND_{j}^{\tau_k} \lor ND_{j'}^{\tau_{k-1}} \nsim ND_{j}^{\tau_k})\}$
%\end{equation}
contain all neutrons incident to neutron $ND_{j}^{\tau_k}$.
Finally for the $j$-th electron of dimension $\tau_k$ we let
%\begin{equation}
$\mathcal{N}_{ee_{j}^{\tau_k}} := \{ ee_{j'}^{\tau_{k-1}} ~|~ ee_{j'}^{\tau_{k-1}} \prec ee_{j}^{\tau_k} \land (ee_{j'}^{\tau_{k-1}} \sim ee_{j}^{\tau_k} \lor ee_{j'}^{\tau_{k-1}} \nsim ee_{j}^{\tau_k}) \}$
%\end{equation}
contain all electrons incident to $ee_{j}^{\tau_k}$. For the $i$-th atom of dimension $k$ we similarly let
%\begin{equation}
$\mathcal{N}_{A_{j}^{k}} := \{ A_{j'}^{k-1} ~|~ A_{j'}^{k-1} \prec A_{j}^{k} \land (A_{j'}^{k-1} \sim A_{j}^{k} \lor A_{j'}^{k-1} \nsim A_{j}^{k})\}$
%\end{equation}
contain all atoms incident to $A_{j}^{k}$. The set $\mathcal{N}_{A_{j}^{k}}$ then describes an atom's local environment. One can choose radius $r > 0$, and distance function $d$ such that:
\begin{equation}
\mathbf{ENV}_{A_{\ell}^{k}} := \{A_{i}^{k-1} ~|~ \forall (i,j)~d(A_{i}^{k-1}, A_{j}^{k-1}) \leq r \land A_{i}^{k-1} \prec A_{\ell}^{k} \land A_{j}^{k-1} \prec A_{\ell}^{k}\}.
\end{equation}
$\mathbf{ENV}_{A_{\ell}^{k}}$ more generally defines environments in a molecule or polyatomic system.~\\
This setup enables us to write out an equation analogous to (1). In particular, let $X$ be a polyatomic complex. Let $d_{n, k}: A_{i_1}^{k} \times A_{i_2}^{k} \times \cdots \times A_{i_n}^{k} \to \mathbb{R}$  be a valid metric defined on products of atoms (e.g. $\sup$ metric). Then let $\mathcal{V}_{bond} := \{(A_{i}^{k}, A_{j}^{k}) ~|~ d_{2, k}(A_{i}^{k}, A_{j}^{k}) < r_{bond}~ \forall k \leq \dim(X)\}$. In essence, $\mathcal{V}_{bond}$ is the set of all pairs of atoms whose separation is only $1$ bond, given by $r_{bond}$. %This definition, namely the choice to define bond parameters over pairs of atoms, follows the bond parameterization of the General Amber Force Field \citep{wangAMBERFF}. 
Let $\mathcal{V}_{angle} := \{(A_{i}^{k}, A_{j}^{k}, A_{\ell}^{k}) ~|~ d_{3, k}(A_{i}^{k}, A_{j}^{k}, A_{\ell}^{k}) < r_{angle}~ \forall k \leq \dim(X)\}$. In essence, $\mathcal{V}_{angle}$ is the set of all triples of atoms whose separation is only $2$ bonds, thereby forming an angle, given by $r_{angle}$. %This definition, namely the choice to define angle parameters over triples of atoms, follows the angle parameterization of the General Amber Force Field \citep{wangAMBERFF}. 
Let $\mathcal{V}_{dihedral} := \{(A_{i}^{k}, A_{j}^{k}, A_{\ell}^{k}, A_{m}^{k}) ~|~ d_{4, k}(A_{i}^{k}, A_{j}^{k}, A_{\ell}^{k}, A_{m}^{k}) < r_{dih}~ \forall k \leq \dim(X)\}$. In essence, $\mathcal{V}_{dihedral}$ is the set of all quadruples of atoms whose separation is only $3$ bonds, thereby forming a dihedral potential or torsion potential, given by $r_{dih}$. 
%This definition, namely the choice to define torsion parameters over quadruples of atoms, follows the torsional angle parameterization of the General Amber Force Field \citep{wangAMBERFF}. 
Let $\mathcal{V}_{nb} := \{(A_{i}^{k} \times A_{j}^{k} ) ~|~ d_{2, k}(A_{i}^{k} \times A_{j}^{k}) > r_{nb}~ \forall k \leq \dim(X)\}$. In essence, $\mathcal{V}_{nb}$ is the set of all tuples of atoms whose separation is greater than $3$ bonds given by $r_{nb}$. 
The definitions of $\mathcal{V}_{bond}$, $\mathcal{V}_{angle}$, $\mathcal{V}_{dihedral}$, and $\mathcal{V}_{nb}$ are motivated by the parameterizations given by the General Amber Force Field \citep{wangAMBERFF, lin2019force}. For classification one can apply numerous force fields such as GAFF \citep{QMDFF, ReaxFF, wangAMBERFF}. The function $\rho$ describes the separation between the atoms. The function $\vartheta$ determines the angle between atoms. The function $\xi$ determines angles formed by the planes defined by atoms.
\begin{multline}
V_{total}
= \sum_{v_r \in \mathcal{V}_{bond}} k_{r}(\rho(v_r) - r_{eq})^{2} + \sum_{v_a \in \mathcal{V}_{angle}} k_{\theta}(\vartheta(v_a) - \theta_{eq})^{2}
\\+ \sum_{v_d \in \mathcal{V}_{dihedral}} k_{t}[1+ \cos(n \cdot \xi(v_d) - \gamma)] + \sum_{v_b \in \mathcal{V}_{nb}} \left[ \frac{A_{ij}}{\rho(v_b)^{12}} - \frac{B_{ij}}{\rho(v_b)^{6}} \right] + \frac{q_i q_j}{4 \pi \epsilon \rho(v_b)}
\end{multline}
More complex interactions can be modeled by considering sets like
\begin{equation}
\mathbf{Int}_{n} := \{(A_{i_1}^{k} \times \cdots \times A_{i_n}^{k} ) ~|~ d_{n, k}(A_{i_1}^{k} \times \cdots \times A_{i_n}^{k}) < r~ \forall k \leq \dim(X)\}
\end{equation}
Note that all parameters can be chosen to reproduce experimental data or quantum mechanical calculations \citep{QMDFF, wangAMBERFF}.
\subsection{Compute Cost}
\label{appendix:computecost}
All reported experiments can be reproduced using an AWS m7g.4xlarge instance (16 vCPU, 64 GiB) in under three hours per experiment. The full project required more compute than the experiments reported in the paper. This is because we conducted a wider variety of experiments on different datasets which are not reported. Certain experiments require resources equivalent to that of an AWS p5.48xlarge (192 vCPU, 2TiB, 8 GPU-H100) and approximately ten hours per experiment.

\end{document}